\documentclass[twocolumn]{autart}    

\makeatletter
\let\algorithm\@undefined
\let\endalgorithm\@undefined
\makeatother
\usepackage{algorithm}

\usepackage[utf8]{inputenc}
\usepackage{graphicx}
\usepackage[version=4]{mhchem}
\usepackage{siunitx}
\usepackage{longtable,tabularx}
\usepackage{pdfsync} 
\usepackage{subcaption} 	
\usepackage{amsmath} 		
\usepackage{amssymb} 	  	
\usepackage{url}

\usepackage{mathtools}    	
\usepackage{float}        	
\usepackage{enumitem}     	
\usepackage{booktabs}     	
\usepackage{multirow}     	
\usepackage{makecell}     	
\usepackage{bm}           	

\makeatletter
\let\c@defn\relax

\newtheorem{defn}{Definition}
\makeatother

\newtheorem{remark}{Remark}[section]

\begin{document}

\begin{frontmatter}

\title{Recursive Gaussian Process State Space Model\thanksref{footnoteinfo}} 

\thanks[footnoteinfo]{Corresponding author: Lin Cheng (chenglin5580@buaa.edu.cn).}

\author[BUAA,CALT]{Tengjie Zheng}\ead{ZhengTengjie@buaa.edu.cn},
\author[BUAA,CALT]{Haipeng Chen}\ead{Hitchenhp@163.com},
\author[BUAA,CALT]{Lin Cheng\thanksref{footnoteinfo}}\ead{chenglin5580@buaa.edu.cn},
\author[BUAA,CALT]{Shengping Gong}\ead{gongsp@buaa.edu.cn},
\author[BUAA,CALT]{Xu Huang}\ead{xunudt@126.com}

\address[BUAA]{School of Astronautics, Beihang University, Beijing, 102206, China}
\address[CALT]{State Key Laboratory of High-Efficiency Reusable Aerospace Transportation Technology, Beijing, 102206, China}


\begin{abstract}                          
    Learning dynamical models from data is not only fundamental but also holds great promise for advancing principle discovery, time-series prediction, and controller design. Among various approaches, Gaussian Process State-Space Models (GPSSMs) have recently gained significant attention due to their combination of flexibility and interpretability. However, for online learning, the field lacks an efficient method suitable for scenarios where prior information regarding data distribution and model function is limited. To address this issue, this paper proposes a recursive GPSSM method with adaptive capabilities for both operating domains and Gaussian process (GP) hyperparameters. Specifically, first-order linearization is first applied to derive a Bayesian update equation for the joint distribution between the system state and the GP model, enabling closed-form and domain-independent learning. Second, an online selection algorithm for inducing points is developed based on informative criteria to achieve lightweight learning. Third, to support online hyperparameter optimization, historical measurement information is recovered from the current filtering distribution. Comprehensive evaluations on both synthetic and real-world datasets demonstrate the superior accuracy, computational efficiency, and adaptability of the proposed method compared to state-of-the-art online GPSSM techniques.
\end{abstract}

\end{frontmatter}

\section{Introduction}
State-space models (SSMs), also known as hidden Markov models (HMMs), describe the evolution and observation processes of latent states through a state transition model and a measurement model. By summarizing historical information into the latent state, SSMs provide a concise representation of system dynamics and are widely applied in fields such as motion planning \cite{he2025novel}, target tracking \cite{kang2026stochastic,zhang2023group}, and control systems design \cite{breschi2023data,kessler2026design,da2025fault,2025Experience}. Nevertheless, due to the complexity of real-world systems, establishing precise SSMs based on first principles is challenging. To achieve sufficiently accurate system models, extensive research has focused on data-driven SSM modeling \cite{Berntorp2021,rangapuram2018deep,gedon2021deep,frigolaBayesianTimeSeries2015,zheng2025model}.

In the past decade, a novel SSM learning method based on the Gaussian process (GP), a nonparametric modeling technique, has been explored \cite{turner2010state,frigolaBayesianTimeSeries2015}. 
Mathematically, a GP is a probability distribution over the function space, providing a flexible and probabilistic framework for modeling unknown functions.
GPs not only yield predictions but also quantify uncertainty, making them widely used in machine learning \cite{rasmussenGaussianProcessesMachine2008} and system identification \cite{pillonetto2022regularization,care2023kernel}. 
In the context of SSM learning, GPs can be employed to model the transition and measurement 
functions, resulting in what are known as Gaussian process state-space models (GPSSMs) \cite
{frigolaBayesianTimeSeries2015}.
By employing GPs, GPSSMs offer flexible learning for the dynamical system and provide prediction uncertainty, which is crucial for safety-critical applications \cite{fisac2018general}.  
To date, various GPSSM methods have been developed, including sample-based methods \cite{frigola2013bayesian,frigola2014variational,svensson2017flexible,fan2023free} and variational inference (VI)-based methods \cite{eleftheriadis2017identification,doerrProbabilisticRecurrentStateSpace2018,Ialongo2019,lindinger2022laplace,Lin2024}. 
For example, the early variational GPSSM method proposed by Frigola et al. \cite{frigola2014variational} combines variational Bayes and sequential Monte Carlo (SMC) to learn the GPSSM by iteratively updating the GP and the state trajectory. Subsequently, the probabilistic recurrent state space model (PRSSM) \cite{doerrProbabilisticRecurrentStateSpace2018} adopts the training approach of recurrent neural networks (RNNs) to optimize the variational posterior of the initial state and GP. In view of the limitation of the mean-field assumption in PRSSM, the variationally coupled dynamic trajectories (VCDT) \cite{Ialongo2019} parameterize the coupling between the state and inducing points. Recently, the ensemble Kalman filter-aided variational inference (EnVI) \cite{Lin2024} employs the ensemble Kalman filter (EnKF) to capture the coupling between state and GP, eliminating the need for the additional inference network required in previous methods.
These methods have achieved significant progress in GPSSM learning, but most studies focus solely on offline scenarios. 
In engineering practice, many systems experience non-stationary changes or varies cross tasks, and there is the possibility of operating in out-of-distribution settings.
Therefore, this underscores the urgent need for online learning methods for GPSSMs, which is the focus of this paper.

Facilitating online learning for GPSSMs presents significant challenges. In addition to the typical constraints of online learning, such as minimizing memory usage, reducing computational complexity, and preventing catastrophic forgetting, GPSSMs introduce further unique difficulties. These challenges are described as follows.
\textbf{(1) Nonlinearity complicates inference.} In GPSSMs, the system state is included in the input to the GP model, introducing inherent nonlinearity. This nonlinearity is intractable, which prevents exact inference through analytic methods.
\textbf{(2) Nonparametric nature of GP.} For standard Gaussian process regression (GPR), the training complexity scales cubically with the number of samples \cite{rasmussenGaussianProcessesMachine2008}. In an online context, as new data arrives continuously, the computational cost of GPs increases rapidly, rendering inference intractable. 
\textbf{(3) Coupling between the system state and the system model.} Since the state is implicitly measured in general, learning the system model requires estimating the system state. However, state estimation depends on the system model itself. As a result, there is a coupling between the state and the model, which increases the complexity of inference. Theoretically, since both are uncertain, they should be treated as latent variables and inferred simultaneously.
\textbf{(4) Hyperparameter online optimization.} GPSSMs involve several hyperparameters, such as GP kernel parameters, which impact the learning accuracy.  
In online settings, hyperparameter optimization presents two main challenges. First, due to memory efficiency considerations, historical measurements are not retained in online scenarios, depriving hyperparameter optimization of explicit information sources. Secondly, since the GP prior depends on kernel hyperparameters, tuning these parameters requires corresponding adjustments to the posterior distribution.
In summary, the online GPSSM problem is quite challenging, and as a result, relevant research remains limited.


Theoretically, implementing online GPSSMs involves Bayesian inference. Therefore, existing methods primarily rely on Bayesian filtering or online variational inference techniques, which can be classified into three categories:
\begin{enumerate}
    \item Augmented Kalman Filter (AKF)-based method \cite{veiback2019learning,Kullberg2020,Kullberg2021}. This approach can jointly infer the system state and the GP model online, and its implementation involves two steps: first, parameterizing the GP with a limited number of parameters using inducing-point approximation \cite{veiback2019learning,Kullberg2020}, or spectrum approximation \cite{Kullberg2021}; and second, augmenting the GP model parameters into the state space and inferring the augmented state using the extended Kalman filter (EKF). Through the AKF-based framework, this method achieves recursive learning and offers advantages in learning speed and stability. However, there is a key limitation: the parameterization of the GP requires prior knowledge of the operating domain, specifically the location of inducing points or the range of basis functions in spectrum approximation \cite{Kullberg2021}. This drawback limits the approach's ability to adapt to the flexible distribution of online data. In addition, there is no feasible method for hyperparameter optimization for this approach.
    \item Particle Filter (PF)-based method \cite{Berntorp2021,zhaoStreamingVariationalMonte2023,liu2023sequential}. 
    This approach is based on the paradigm of the Rao-Blackwellized Particle Filter \cite{grisetti2007improved}.
    It is similar to the AKF-based method, with the key difference being the use of the particle filter (PF) for the inference process. Since PF is inefficient at handling high-dimensional states, GP parameters are not augmented into the state space. Instead, each particle is assigned a GP, which is recursively learned using the particle state. This method can theoretically achieve precise inference for arbitrary observational distributions, provided the number of particles is sufficient. However, despite the inherent issues with PF, such as high computational cost and particle degeneracy \cite{doucet2009tutorial}, it also requires prior knowledge of the operating domain. 
    \item Stochastic Variational Inference (SVI)-based method \cite{frigola2014variational,Lin2024,Park2022}. This approach extends the offline variational inference method of GPSSMs by utilizing stochastic optimization techniques. It aims to retain many of the benefits of offline GPSSMs, such as a flexible operating domain and hyperparameter optimization capabilities. However, stochastic optimization assumes that the data subsampling process is uniformly random, which is not the case in most online situations. As a result, the learning process is driven by instantaneous data, leading to slow learning and catastrophic forgetting.
\end{enumerate}
In summary, the existing methods either lack the ability to tune the operating domains and hyperparameters, or they face issues with learning speed and convergence.

To address the aforementioned shortcomings, this paper proposes a recursive learning algorithm for GPSSMs that is capable of online adaptation to the operating domain and GP hyperparameters.
The main contributions can be summarized as follows:  
Firstly, to implement a domain-independent recursive learning algorithm for GPSSMs, a novel Bayesian update equation is derived without pre-parameterizing the GP. In the derivation process, only mild linearization approximations are applied to achieve a closed-form solution, offering advantages in learning speed and stability. Through continuously expanding the inducing-point set, this method can accommodate real-time data distributions.
Secondly, to address the computational cost issue caused by the increasing number of inducing points, a selection algorithm is developed with both addition and removal operations. To minimize accuracy loss, an optimal discarding equation and selection metric for inducing points are derived based on informative criteria.
Thirdly, to alleviate the burden of offline hyperparameter tuning, an online GP hyperparameter optimization method is proposed. In this method, an approximate observation likelihood model is extracted from the obtained posterior distribution. Utilizing this likelihood model as the information source, the GP hyperparameters are optimized, and the posterior distribution can be adjusted as the hyperparameters change. 
Finally, the effectiveness of the proposed method is validated through two synthetic simulations and a real-world dataset. The experimental results demonstrate the superior accuracy, computational efficiency, and adaptability of the proposed method compared to state-of-the-art online GPSSM techniques.

\section{Problem Formulation}

In this section, we first briefly provide the background of Gaussian processes (GPs) and then give a definition of Gaussian process state-space models (GPSSMs) and the associated online learning problem.

\subsection{Gaussian Processes}

Mathematically, a GP is a probability distribution over function space, which has been developed for regression tasks. 
For learning a single-output function $f(\cdot)$, one can first place a GP prior on it:

\begin{equation}
\begin{aligned}
f(\textbf{x}) \sim \mathcal{GP}\left(m_0(\textbf{x}), 
K_0(\textbf{x}, \textbf{x}^\prime;\bm{\theta}) \right)
\end{aligned}
\end{equation} 
where $m_0(\cdot)$ denotes the prior mean function and $K_0(\cdot, \cdot)$ represents the prior covariance function (also known as kernel function), and $\bm{\theta}$ denotes the kernel hyperparameters. As a common setting, we assume zero-mean GP prior, namely, $m_0(\cdot) \equiv 0$. 
Then, given noise-free function values 
\( \bm{f}_{\mathcal{X}} = \{f(\textbf x_i)\}_{i=1}^{n_\mathcal{X}} \) 
at inputs \( \mathcal{X} = \{\mathbf{x}_i\}_{i=1}^{n_\mathcal{X}}\), 
the posterior distribution for test function values 
\( \bm{f}_* = \{f(\textbf x_i^*)\}_{i=1}^{n_{\mathcal{X}^*}}  \) 
at inputs \( \mathcal{X}^* = \{\mathbf{x}_i^*\}_{i=1}^{n_{\mathcal{X}^*}}\) is \cite{rasmussenGaussianProcessesMachine2008}:

\begin{equation}\label{eq:GP_cond}
\begin{aligned}
p(\bm{f}_*|\bm{f}_{\mathcal{X}}) = \mathcal{N}(
\bm{K}_{*\mathcal{X}} 
\bm{K}_{\mathcal{XX}}^{-1} \bm{f}_{\mathcal{X}}, 
 \bm{K}_{**} 
- \bm{K}_{*\mathcal{X}} \bm{K}_{\mathcal{XX}}^{-1} 
\bm{K}_{\mathcal{X}*} )
\end{aligned}
\end{equation}
Here, we use shorthand notation for the kernel matrix, e.g., \( \bm{K}_{**} \) denotes the auto-covariance of \( \bm{f}_* \), and \( \bm{K}_{* \mathcal{X}} \) the cross-covariance between \( \bm{f}_* \) and \( \bm{f}_{\mathcal{X}} \). This notation convention is consistently used throughout the entire text. 
The above discussion is based on single-output GPs. However, GPs can also be applied to learn multi-output functions by employing a matrix-valued kernel. 
However, for simplicity and clarity, the proposed method is elaborated first based on the single-output GP, and the extension to the multi-output case is discussed in Section \ref{subsec:multi_output}. Hence, in the following, the function to learn will be denoted as $f(\cdot)$ without boldface.

\subsection{Online Gaussian Process State-Space Models}

In this paper, we consider the following discrete-time state-space model (SSM):

\begin{equation}\label{eq:SSM}
\begin{aligned}
\bm{x}_{t+1} &= \bm{F}(\bm{x}_t, f(\bm{x}_t)) + \bm{\omega}_p, 
&\bm{\omega}_p \sim \mathcal N(\bm{0}, \bm{\Sigma}_p)\\
\bm{y}_t &= \bm{g}(\bm{x}_t) + \bm{\omega}_m, 
&\bm{\omega}_m \sim \mathcal N(\bm{0}, \bm{\Sigma}_m)
\end{aligned}
\end{equation}
where the vector $\bm{x} \in \mathbb{R}^{d_x}$ and $\bm{y} \in \mathbb{R}^{d_y}$ respectively denote the system state and measurement. The vectors $\bm{\omega}_p$ and $\bm{\omega}_m$ represent the process and measurement noise, with covariances $\bm{\Sigma}_p$ and $\bm{\Sigma}_m$, respectively. 
Additionally, the function $\bm{F}: \mathbb{R}^{d_x} \times \mathbb{R} \to \mathbb{R}^{d_x}$ denotes a known function structure for the transition model\footnote{
   This structure of \( \bm F(\cdot) \) can represent various models of interest, such as when only some dimensions of the transition model are unknown, or a discretized continuous-time transition model \( \bm F(\bm x_t, \bm f(\bm x_t)) = \bm x_t + \bm f(\bm x_t) \Delta t \), where \( \Delta t \) is the time step and the unknown component is the time-derivative model \( \dot{\bm x} = \bm f(\bm x) \). Through using a varying time step \( \Delta t \), the latter can be used for irregular measurement scenarios commonly occurring in biochemical sciences.
}, which contains an unknown component, represented as the function $f: \mathbb{R}^{d_x} \to \mathbb{R}$. To ensure identifiability \cite{Lin2024}, the measurement model $\bm{g}: \mathbb{R}^{d_x} \to \mathbb{R}^{d_y}$ is assumed to be known\footnote{
    Even if the measurement model is unknown in practice, we can augment the measurement $\bm{y}$ into the state space and transform the problem into one with a known measurement model \cite{frigolaBayesianTimeSeries2015}.
}. 

To learn the unknown SSMs, GPSSMs place GP priors on the unknown function $f(\cdot)$, resulting in the following model \cite{frigolaBayesianTimeSeries2015}:

\begin{subequations}\label{eq:GPSSM}
\begin{gather}
\bm{x}_0 \sim p(\bm{x}_0)  \\
f(\cdot) \sim \mathcal{GP}(\bm 0, K_0(\cdot, \cdot;\bm{\theta}))  \\ 
\label{eq:tran_density}
\bm{x}_{t+1} | \bm{x}_{t},\bm{f} \sim 
\mathcal N \left(\bm{x}_{t+1} \, \big| \,  \bm{F}(\bm{x}_t, f(\bm{x}_t)), \, \bm{\Sigma}_p \right) \\
\label{eq:meas_density}
\bm{y}_t | \bm{x}_t \sim 
\mathcal N \left(\bm{y}_t \, \big| \,  \bm{g}(\bm{x}_t), \, \bm{\Sigma}_m \right)
\end{gather}
\end{subequations}
where $p(\bm{x}_0) = \mathcal N(\bm x_0|\bm\mu_0, \bm P_0)$ denotes a prior Gaussian distribution for the initial state $\bm{x}_0$, and we use the notation $\bm{f} = \{f(\bm x) : \bm x \in \mathbb R^{d_x}\}$ to denote all function values of the function $f(\cdot)$. For conciseness, we omit the control input $\bm c \in \mathbb R^{d_c}$ from the transition and measurement models, which can be easily incorporated by augmenting it into the input of these models and the function $f(\cdot)$.

In the online setting, measurements \( \bm{y} \) arrive sequentially, and we assume that historical measurements are not retained for computational and storage reasons. Given this streaming data setting, the inference task of online GPSSMs is to sequentially approximate the filtering distribution \( p(\bm{x}_t, \bm{f}|\bm{y}_{1:t}) \), using the previous result \( p(\bm{x}_{t-1}, \bm{f}|\bm{y}_{1:t-1}) \) and the current measurement \( \bm{y}_t \). 
As illustrated in the introduction, several challenges exist in this task, including nonlinearity, coupling between state and GP, as well as the nonparametric nature of GPs and the hyperparameter optimization problem. To address these difficulties, a novel online GPSSM approach will be developed in the next two sections, with the ability to adapt to both operating domains and hyperparameters.

\section{Bayesian Update Equation for Online GPSSMs}\label{sec:update_eq}

This section derives a Bayesian update equation for online GPSSMs with minimal approximations to achieve recursive learning in an arbitrary operating domain.

\subsection{Two-Step Inference}

The online inference of GPSSMs is essentially a filtering problem for the hidden Markov model; therefore, most methods are based on filtering techniques, such as the extended Kalman filter (EKF) and particle filter (PF). 
Considering that the PF not only exacerbates the computational cost of the GP but also suffers from particle degeneracy, we adopt the EKF in this paper.
The widely adopted EKF is a two-step version (one-step version can be seen in \cite{ljung2003asymptotic}), which is more numerically stable \cite{oconnellNeuralFlyEnablesRapid2022} and can handle irregular measurements.
In each update, the two-step EKF propagates the state distribution and then corrects it with the current measurement, which are known as the prediction (time update) and correction (measurement update) steps, respectively. In the context of GPSSMs, considering the coupling between the system state $\bm x$ and the GP, the distribution to be propagated and corrected is the joint distribution of both. These two operations can be expressed by:

\begin{align}
\label{eq:pred_step}
&p(\bm{x}_{t+1}, \bm{f}|\bm{y}_{1:t}) = 
\int  p(\bm{x}_{t+1}|\bm{x}_t, \bm{f}) 
p(\bm{x}_{t}, \bm{f}|\bm{y}_{1:t}) \mathrm{d}\bm{x}_t \\
\label{eq:corr_step}
&p(\bm{x}_{t+1}, \bm{f}|\bm{y}_{1:t+1}) = \dfrac{p(\bm{y}_{t+1}|\bm{x}_{t+1}) 
p(\bm{x}_{t+1}, \bm{f}|\bm{y}_{1:t})}{p(\bm{y}_{t+1}|\bm{y}_{1:t})} 
\end{align}
where \( p(\bm{x}_{t+1}|\bm{x}_t, \bm{f}) \) and \( p(\bm{y}_{t+1}|\bm{x}_{t+1}) \) are the transition and measurement densities as defined in \eqref{eq:GPSSM}.
Additionally, \( p(\bm{x}_t, \bm{f} | \bm{y}_{1:t}) \) denotes the joint posterior at time \( t \), and \( p(\bm{x}_{t+1}, \bm{f} | \bm{y}_{1:t}) \) is the joint prior or predicted distribution at time \( t+1 \).  
To evaluate the distribution in equations \eqref{eq:pred_step} and \eqref{eq:corr_step}, the main challenges are the nonparametric nature of the GP (that is, $\bm f$ is infinite dimensional) and the nonlinearity in the transition and measurement models.  
Existing methods \cite{Berntorp2021,veiback2019learning,Kullberg2020,Kullberg2021,zhaoStreamingVariationalMonte2023} typically first address the former by pre-parameterizing the GP into a fixed structure and then applying nonlinear filtering to handle the latter. 
However, the pre-parameterization restricts the operating domain of the model (e.g., learning is limited to a predefined range such as $[-1, 1]$ in the one-dimensional input domain), reducing the flexibility of the algorithm.
To overcome this, we reverse the process: first handling the nonlinearity without any GP approximation, then addressing the nonparametric nature. We achieve the first procedure in the next subsection, resulting in an update equation with a factorized approximate distribution.

\subsection{Factorized Approximate Distribution}\label{subsec:factor_dist}

To address the nonlinearity within the transition and measurement models, the following theorem is derived in this work by applying first-order linearization. 

\begin{defn}[First-order linearization]\label{defn1}
    Consider a Gaussian probabilistic model
    \[
    p(\bm a \mid \bm b) = \mathcal{N}\big(\bm a \mid \bm h(\bm b),\, \bm S\big),
    \]
    where $\bm h(\cdot)$ is a differentiable mapping and $\bm S$ is the covariance matrix. 
    The first-order linearization of this model around the mean $\bm m_b$ is defined as
    \[
    p(\bm a \mid \bm b) \approx 
    \mathcal{N}\!\left(\bm a \,\middle|\, \bm h(\bm m_b) + 
    \frac{\partial \bm h}{\partial \bm b}\bigg|_{\bm m_b} (\bm b - \bm m_b),\, \bm S\right).
    \]
\end{defn}

\begin{thm}\label{theorem1}
Let $q(\bm{x}_t, \bm{f})$ denote an approximation to the joint distribution $p(\bm{x}_t, \bm{f})$ obtained in the prediction step \eqref{eq:pred_step} and the correction step \eqref{eq:corr_step}, where, for brevity, $p(\bm{x}_t, \bm{f})$ represents either $p(\bm{x}_{t+1}, \bm{f} \mid \bm{y}_{1:t})$ or $p(\bm{x}_{t+1}, \bm{f} \mid \bm{y}_{1:t+1})$.
Let $\bm{u} \triangleq \{ u_i \}_{i=1}^{n_u}$ denote a finite set of inducing points, where each inducing point $u_i = f(\bm{x}_i)$ is the function value at the input location $\bm{x}_i$. 
The set $\bm u$ can be initialized as an arbitrary finite set or as the empty set.

In each prediction step and correction step, by applying the first-order linearization described in Definition \ref{defn1} to both the transition model \eqref{eq:tran_density} and the measurement model \eqref{eq:meas_density}, and by augmenting the set of inducing points $\bm{u}$ with the function value $\bm{f}(\mathbb{E}_{q(\bm{x}_t, \bm{u})}[\bm{x}_t])$, the approximate distribution $q(\bm{x}_t, \bm{f})$ takes the following factorized form:
    \begin{equation}\label{eq:approx_dist0}
        q(\bm{x}_t, \bm{f}) = p(\bm{f}_{\ne u}|\bm{u}) q(\bm{x}_t, \bm{u})
    \end{equation}
where $\bm f_{\ne u} = \bm{f} \setminus \bm{u}$ denotes all function values except $\bm u$, and $p(\bm{f}_{\ne u}|\bm{u})$ denotes the GP prior conditional, analogous to \eqref{eq:GP_cond}:
    \begin{equation}\label{eq:cond_prior}
    \begin{aligned}
    &p(\bm{f}_{\ne u}|\bm{u}) \\
    &= \mathcal N \left( 
        \bm{K}_{f_{\ne u}u} \bm{K}_{uu}^{-1} \bm{u}, \,
        \bm{K}_{f_{\ne u}f_{\ne u}} - \bm{K}_{f_{\ne u}u} \bm{K}_{uu}^{-1} \bm{K}_{uf_{\ne u}}
    \right)
    \end{aligned}
    \end{equation}
Furthermore, $q(\bm{x}_t, \bm{u})$ in \eqref{eq:approx_dist0} is a Gaussian distribution.
\end{thm}

Given Theorem \ref{theorem1}, since $p(\bm{f}_{\ne u}|\bm{u})$ is known, we only need to update the finite-dimensional distribution $q(\bm{x}_t, \bm{u})$, which build a foundation for addressing the non-parametric nature of the GP. Moreover, the approximate joint distribution $q(\bm{x}_t, \bm{f})$ remains Gaussian, which can be evaluated using closed-form update equations. In the following, we will prove Theorem \ref{theorem1} and then derive these equations.

\begin{pf}
To facilitate the proof, we first introduce some notation regarding the approximate distributions \( q(\bm{x}_t, \bm{u}) \) and \( q(\bm{x}_t, \bm{f}) \). Specifically, these two distributions are expressed as:

\begin{equation}\label{eq:approx_dist}
\begin{aligned}
q(\bm{x}_t, \bm{u}) & = \mathcal{N}\left(
\begin{bmatrix} \bm{x}_t \\ \bm{u} \\ \end{bmatrix} \Bigg |
\begin{bmatrix} \bm{\mu}_t \\ \bm{m}_{u} \\ \end{bmatrix},
\begin{bmatrix}
\bm{P}_t & \bm{V}_{xu} \\ 
\bm{V}_{xu}^T & \bm{S}_{uu} \\ 
\end{bmatrix} \right) \\ 
q(\bm{x}_t, \bm{f}) & = \mathcal{N}\left(
\begin{bmatrix} \bm{x}_t \\ \bm{f} \\ \end{bmatrix} \Bigg |
\begin{bmatrix} \bm{\mu}_t \\ \bm{m}_{f} \\ \end{bmatrix},
\begin{bmatrix}
\bm{P}_t & \bm{V}_{xf} \\ 
\bm{V}_{xf}^T & \bm{S}_{ff} \\ 
\end{bmatrix} \right)
\end{aligned}
\end{equation}
In this expression, the moments of the joint distribution are distinguished based on the state \( \bm{x}_t \) and inducing points \( \bm{u} \) or function values \( \bm{f} \). For \( q(\bm{x}_t, \bm{u}) \), 
the vectors \( \bm{\mu}_t = \mathbb{E}_{q(\bm{x}_t, \bm{u})}[\bm x_t]  \) 
and \( \bm{m}_{u} = \mathbb{E}_{q(\bm{x}_t, \bm{u})}[\bm u] \) represent the means of the state \( \bm{x}_t \) and the inducing points \( \bm{u} \), respectively. 
\( \bm{P}_t = \operatorname{Var}_{q(\bm{x}_t, \bm{u})}[\bm x_t] \) denotes the autocovariance of \( \bm{x}_t \), 
while \( \bm{V}_{xu} = \operatorname{Cov}_{q(\bm{x}_t, \bm{u})}[\bm x_t, \bm u] \) represents the cross-covariance between \( \bm{x}_t \) and \( \bm{u} \), 
and \( \bm{S}_{uu} = \operatorname{Var}_{q(\bm{x}_t, \bm{u})}[\bm u]\) signifies the autocovariance of \( \bm{u} \). 
For the moments of \( q(\bm{x}_t, \bm{f}) \), the notations follow the same logic, and the values are fully specified by the moments of \( q(\bm{x}_t, \bm{u}) \) given \eqref{eq:approx_dist0}, specifically:

\begin{equation}\label{eq:qf}
\begin{aligned}
&\bm{m}_f = \bm{k}_{fu} \bm{m}_u \\
&\bm{V}_{xf} = \bm{V}_{xu} \bm{k}_{fu}^T \\
&\bm{S}_{ff} = \bm{K}_{ff} + \bm{k}_{fu} 
\left( \bm{S}_{uu} - \bm{K}_{uu} \right)
\bm{k}_{fu}^T
\end{aligned}
\end{equation}
where we use the  shorthand notation \( \bm{k}_{fu} = \bm{K}_{fu} \bm{K}_{uu}^{-1} \). 
This equation can be easily derived by using \eqref{eq:approx_dist0} and the conditional distribution \( p(\bm{f}_{\ne u}|\bm{u}) \) \eqref{eq:cond_prior}.

With the notation introduced above, the proof of Theorem \ref{theorem1} is performed using mathematical induction.
First, it can be demonstrated that at time \( 0 \), the joint distribution naturally satisfies the factorized form \eqref{eq:approx_dist0}:

\begin{equation}
\begin{aligned}
q(\bm{x}_0, \bm{f}) & \triangleq p(\bm{x}_0, \bm{f}) \\
&= p(\bm{f})p(\bm{x}_0) \\
&= p(\bm{f}_{\ne u}|\bm{u})  p(\bm{u}) p(\bm{x}_0) \\
&= p(\bm{f}_{\ne u}|\bm{u}) q(\bm{x}_0, \bm{u})
\end{aligned}
\end{equation}
where $q(\bm{x}_0, \bm{u}) \triangleq p(\bm{u}) p(\bm{x}_0)$ is Gaussian, $p(\bm{f})$ denotes the GP prior over all function values $\bm{f}$, and \( \bm{u} \) can be any collection of function values or an empty set. 
Therefore, the next step is to prove that, assuming after the update at time $t$ the approximate joint distribution $q(\bm x_t, \bm f)$ satisfies the factorized form \eqref{eq:approx_dist0}, the approximate updates in the prediction step \eqref{eq:pred_step} and the correction step \eqref{eq:corr_step} at time $t+1$ can also preserve the factorized form \eqref{eq:approx_dist0}. This is conducted in the following.

For the prediction step, to overcome the nonlinearity, the transition model \eqref{eq:tran_density} is linearized around the mean of the state $\bm x_t$ and the GP as follows:

\begin{equation}\label{eq:linearized_model}
\begin{aligned}
    &p(\bm{x}_{t+1} | \bm{x}_{t}, \bm{f}) \\
    &\approx \mathcal{N}\Big(\bm{x}_{t+1}\Big| \bm{F}_t + \bm{A}_x (\bm{x}_t - \bm{\mu}_t) 
        + \bm{A}_f (f_t - m_{f_t}), \bm{\Sigma}_p \Big) \\
    &\triangleq q(\bm{x}_{t+1} | \bm{x}_t, f_t).
\end{aligned}
\end{equation}
where we use the notation for moments in \eqref{eq:approx_dist} and the shorthand notation 
\( f_t = f(\bm{\mu}_t) \), $m_{f_t} = \mathbb E_{q(\bm{x}_t, \bm{f})}[f_t]$, \( \bm{F}_t = \bm{F}(\bm{\mu}_t, m_{f_t}) \). Additionally, the two Jacobian matrices in the equation are:
 
\begin{equation}
\begin{aligned}
\bm{A}_f &= \frac{\partial \bm{F}(\bm{\mu}_t, f)}{\partial f} \bigg|_{m_{f_t}} \\
\bm{A}_x &= \frac{\partial \bm{F}(\bm{x}, m_{f(\bm{x})})}{\partial \bm{x}} \bigg|_{\bm{\mu}_t}
= \frac{\partial \bm{F}(\bm{x}, m_{f_t})}{\partial \bm{x}} \bigg|_{\bm{\mu}_t}
+ \bm{A}_f \frac{\partial m_{f(\bm{x})}}{\partial \bm{x}} \bigg|_{\bm{\mu}_t}
\end{aligned}
\end{equation}
where $\bm A_f$ is the Jacobian matrix of the transition function $\bm{F}$ with respect to the function value $f$ evaluated at $m_{f_t}$, and $\bm A_x$ is the Jacobian matrix of $\bm{F}$ with respect to the state $\bm{x}$ evaluated at $\bm{\mu}_t$, which includes both the direct effect of $\bm{x}$ on $\bm{F}$ and the indirect effect through the function value $f(\bm{x})$.


As shown in \eqref{eq:linearized_model}, the linearized transition model  \( q(\bm{x}_{t+1} | \bm{x}_t, f_t) \) depends solely on the function value \( f_t \). 
Therefore, if we include \( f_t \) in the inducing-point set, that is, \( \bar{\bm{u}} = [\bm{u}^T, f_t^T]^T \), and use the linearized model in \eqref{eq:linearized_model}, the prediction result in \eqref{eq:pred_step} keeps the factorized form:

\begin{equation}\label{eq:approx_prior}
\begin{aligned}
q^-(\bm{x}_{t+1}, \bm{f})
&= \int q(\bm{x}_t, \bm{f}) q(\bm{x}_{t+1}|\bm{x}_t, f_t)
\mathrm{d} \bm{x}_t \\
&= \int p(\bm{f}_{\ne \bar u}|\bar {\bm{u}}) q(\bm{x}_t, \bar {\bm{u}}) q(\bm{x}_{t+1}|\bm{x}_t, f_t)
\mathrm{d} \bm{x}_t \\
&= p(\bm{f}_{\ne \bar u}|\bar {\bm{u}}) \int q(\bm{x}_t, \bar {\bm{u}}) q(\bm{x}_{t+1}|\bm{x}_t, f_t)
\mathrm{d} \bm{x}_t \\
&= p(\bm{f}_{\ne \bar u}|\bar {\bm{u}}) q^-(\bm{x}_{t+1}, \bar {\bm{u}})
\end{aligned}
\end{equation}
where we use the superscript "-" to denote predicted distributions, and $q^-(\bm{x}_{t+1}, \bar {\bm{u}}) = 
\int q(\bm{x}_t, \bar {\bm{u}}) q(\bm{x}_{t+1}|\bm{x}_t, f_t)
\mathrm{d} \bm{x}_t$ is Gaussian. It can be observed that the factorized form is preserved at the cost of augmenting the inducing-point set, an inevitable consequence of the non-parametric nature of the GP. 

For the correction step, the handling is similar.
Specifically, linearize the measurement model \eqref{eq:meas_density} as follows:


\begin{equation}
\begin{aligned}
&p(\bm{y}_{t+1}|\bm{x}_{t+1}) \\
&\approx \mathcal{N}\left(\bm{y}_{t+1} \bigg|
\bm{g}(\bm{\mu}_{t+1}^-) + \bm{C}_x(\bm{x}_{t+1} - \bm{\mu}_{t+1}^-), \bm{\Sigma}_m \right) \\
&\triangleq q(\bm{y}_{t+1}|\bm{x}_{t+1})
\end{aligned}
\end{equation}
where $\bm{\mu}_{t+1}^- = \mathbb E_{q^-(\bm{x}_{t+1}, \bar {\bm{u}})}[\bm x_{t+1}]$ denotes the predicted mean of state $\bm{x}_{t+1}$ and the measurement Jacobian is $\bm{C}_x = \partial \bm{g}(\bm{x})/\partial \bm{x} \big|_{\bm{\mu}_{t+1}^-}$. By incorporating the linearized measurement model $q(\bm{y}_{t+1}|\bm{x}_{t+1})$ into the correction step \eqref{eq:corr_step}, the factorized form can be retained:

\begin{equation}\label{eq:approx_post}
\begin{aligned}
q(\bm{x}_{t+1}, \bm{f}) &\propto q^-(\bm{x}_{t+1}, \bm{f}) q(\bm{y}_{t+1}|\bm{x}_{t+1}) \\
&= p(\bm{f}_{\ne \bar u}|\bar{\bm{u}}) q^-(\bm{x}_{t+1}, \bar{\bm{u}}) q(\bm{y}_{t+1}|\bm{x}_{t+1}) \\
&\propto p(\bm{f}_{\ne \bar u}|\bar{\bm{u}}) q(\bm{x}_{t+1}, \bar{\bm{u}})
\end{aligned}
\end{equation}
where $q(\bm{x}_{t+1}, \bar{\bm{u}}) \propto q^-(\bm{x}_{t+1}, \bar{\bm{u}}) q(\bm{y}_{t+1}|\bm{x}_{t+1})$ is Gaussian.

\end{pf}

Thus far, we have proved Theorem \ref{theorem1} using mathematical induction. In this proof, we only use first-order linearization, and the inducing-point set is continuously expanded to match the data distribution.
In contrast to existing methods \cite{Berntorp2021,veiback2019learning,Kullberg2020,Kullberg2021,zhaoStreamingVariationalMonte2023}, this result does not rely on pre-parameterizing the GP and thereby can learn over an arbitrary operating domain.
Furthermore, in the derivation, we also obtain the approximate prediction equation \eqref{eq:approx_prior} and correction equation \eqref{eq:approx_post} for the joint distribution \( q(\bm{x}_t, \bm{f}) \), which can be used to derive practical update equations for it. It is evident that the finite dimensional distribution \( q(\bm{x}_t, \bm{u}) \) contains the essential information of \( q(\bm{x}_t, \bm{f}) \) and thus is the only component that requires updating. Given that \( q(\bm{x}_t, \bm{u}) \) is Gaussian, we only need to derive its moment matching equations, which are presented in the next subsection.

\subsection{Moment Matching Equation}

For simplicity in the moment matching equation, we define the union of the system state $\bm x_t$ and inducing points $\bm u$ as an augmented state, i.e., \( \bm{X}_t = [\bm{x}_t^T, \bm{u}^T]^T \). Consequently, the approximate distribution \( q(\bm{x}_t, \bm{u}) \) can be expressed as \( q(\bm{X}_t) \), with its mean and covariance denoted as \( \bm{\xi}_t = \mathbb E_{q(\bm X_t)}[\bm X_t] \) and \( \bm{\Sigma}_t = \mathrm{Var}_{q(\bm X_t)}[\bm X_t]\), respectively. Based on this definition, the moment matching equation becomes a recursive equation for \( \bm{\xi}_t \) and \( \bm{\Sigma}_t \).

For the prediction step, given the update equation \eqref{eq:approx_prior}, it includes two operations: augmenting the inducing-point set and propagating the moments.
Therefore, we should first add the point \( f_t \) into the inducing-point set $\bm u$, resulting into the new set $\bar{\bm u}$.
For brevity, we define the new augmented state \( \bar{\bm{X}}_t = [\bm{x}_t^T, \bar{\bm{u}}^T]^T \), similar to \( \bm{X}_t \). Correspondingly, we have the approximate distribution \( q(\bar{\bm{X}}_t) \) with mean \( \bar{\bm{\xi}}_t \) and covariance \( \bar{\bm{\Sigma}}_t \), which can be evaluated using \eqref{eq:approx_dist} and \eqref{eq:qf} by letting \( \bm{f} = \bar{\bm{u}} \). Then, by applying the update equation \eqref{eq:approx_prior}, the moment matching equation for the prediction step can be easily derived:

\begin{equation}\label{eq:pred}
\begin{aligned}
&\bar{\bm{\xi}}_{t+1}^- = [\bm{F}_t^T, \bm{m}_{\bar{u}}^T]^T \\
&\bar{\bm{\Sigma}}_{t+1}^- = \bm{\Phi} \bar{\bm{\Sigma}}_t \bm{\Phi}^T 
+ \bm{\Sigma}_{p, \bar{X}} \\
&\bm{\Phi} = \begin{bmatrix}
\bm{A}_x & \bm{0} & \bm{A}_f \\
\bm{0} & \bm{I}_{n_u} & \bm{0} \\
\bm{0} & \bm{0} & 1 \\
\end{bmatrix} \\
& \bm{\Sigma}_{p, \bar{X}} = \begin{bmatrix}
\bm{\Sigma}_p & \bm{0} \\
\bm{0} & \bm{0} \\
\end{bmatrix}
\end{aligned}
\end{equation}
where $\bm F_t$ has been defined in \eqref{eq:linearized_model}, \( \bar{\bm{\xi}}_{t+1}^- \) and \( \bar{\bm{\Sigma}}_{t+1}^- \) represent the predicted mean and covariance for the augmented state \( \bar{\bm{X}}_{t+1} \). \( \bm{\Phi} \) denotes the transition Jacobian matrix for $\bar{\bm{X}}_{t+1}$, \( \bm{\Sigma}_{p, \bar{X}} \) is the process noise covariance for $\bar{\bm{X}}_{t+1}$, and $\bm{I}_{n_u}$ is the identity matrix.

For the correction step, the moment matching equation can be directly obtained from the approximate correction equation in \eqref{eq:approx_post}, which takes the standard EKF form:

\begin{equation}\label{eq:corr}
\begin{aligned}
&\bm{H} = \begin{bmatrix}
\bm{C}_x & \bm{0} \\
\end{bmatrix}\\
&\bm{G} = \bar{\bm{\Sigma}}_{t+1}^- \bm{H}^T \left( \bm{C}_x \bm{P}_{t+1}^- \bm{C}_x^T + \bm{\Sigma}_m \right)^{-1} \\
&\bar{\bm{\xi}}_{t+1} = \bar{\bm{\xi}}_{t+1}^- + \bm{G} \left[ \bm{y}_{t+1} - \bm{g}(\bm{\mu}_{t+1}^-) \right] \\
&\bar{\bm{\Sigma}}_{t+1} = \bar{\bm{\Sigma}}_{t+1}^- - \bm{G} \bm{H} \bar{\bm{\Sigma}}_{t+1}^-
\end{aligned}
\end{equation}
where $\bm{C}_x = \partial \bm{g}(\bm{x})/\partial \bm{x} \big|_{\bm{\mu}_{t+1}^-}$ is the measurement Jacobian with respect to the state $\bm x_{t+1}$. The matrix \( \bm{P}_{t+1}^- \) indicates the predicted variance of the system state \( \bm{x}_{t+1} \). The vectors \( \bar{\bm{\xi}}_{t+1} \) and \( \bar{\bm{\Sigma}}_{t+1} \) are the posterior mean and covariance for the augmented state \( \bar{\bm{X}}_{t+1} \). The matrix \( \bm{H} \) is the measurement Jacobian with respect to the augmented state $\bar{\bm{X}}_{t+1}$, and \( \bm{G} \) is the Kalman gain. 
Therefore, by applying the moment matching equations \eqref{eq:pred} and \eqref{eq:corr}, closed-form recursive inference for online GPSSMs can be achieved in theory.

In summary, this section presents a two-step Bayesian update framework for online GPSSMs, in which first-order linearization is employed to address the nonlinearity in both the transition and measurement models. By applying linearization and expanding the inducing-point set, the joint distribution maintains a factorized form, which leads directly to the moment matching equations \eqref{eq:pred} and \eqref{eq:corr}. 
Linearization is a key step in this method, providing accurate approximations when the system exhibits mild nonlinearity and the variance of the joint distribution is small. This approximation is similar to that used in the EKF, and some techniques can be transferred to the proposed method, such as adding a positive-definite term to the process noise covariance to enhance robustness. Although first-order linearization may be insufficient in some cases, the above derivation establishes a general framework in which other advanced nonlinear filtering techniques can be incorporated by replacing the EKF-based moment matching.

However, the derived method faces a computational burden problem that prevents practical implementation. Specifically, as discussed in Section \ref{subsec:factor_dist}, the number of inducing points will continuously increase in each prediction step. This increase will cause the dimension of the mean vector $\bm{\xi}_t$ and the covariance matrix $\bm\Sigma_t$ to grow, which in turn increases the computational burden and eventually makes inference intractable. In the next section, an approximation method will be introduced to address this issue.

\section{Adaptation of Inducing points and Hyperparameters}\label{sec:adapt}

While the recursive learning algorithm for GPSSMs has been developed, two issues remain: the computational burden caused by continuously adding inducing points, and the online hyperparameter adaptation problem. In Section \ref{subsec:opt_id}, an inducing-point set adjustment algorithm is introduced to maintain limited computational complexity, and in Section \ref{subsec:opt_hp}, an online hyperparameter adaptation method is derived to improve learning accuracy.

\subsection{Dynamically Adjusting the Inducing-Point Set}\label{subsec:opt_id}

As discussed in Section \ref{sec:update_eq}, the computational cost of the learning algorithm increases with the addition of inducing points. This challenge is also encountered in the development of kernel-based online regression algorithms, such as Kernel Recursive Least Squares (KRLS) \cite{VanVaerenbergh2012} and Sparse Online Gaussian Processes (SOGP) \cite{csatoSparseLineGaussian2002,csatoGaussianProcessesIterative2002}. 
These two methods are equivalent to some extent and control computational cost by limiting the size of the inducing-point set to a predefined budget \( M \). 
Specifically, this strategy is achieved through two operations: discarding the least important point when the set exceeds the budget \( M \), and adding only sufficiently novel points.
To implement the discarding and adding operations, the following two key problems must be addressed: (1) how to remove a point with minimal accuracy loss, and (2) how to evaluate the importance or novelty of a point. In the following, we will address these two problems in the context of GPSSMs and develop an adjustment method for the inducing-point set, thereby overcoming the computational burden issue.
For convenience, let \( u_{\mathrm{d}} \) denote the inducing point to be discarded, with index \( i_{\mathrm{d}} \) in the point set, and let \( \bm{u}_{\mathrm{l}} \) represent the inducing points left.

\textbf{Optimal deletion for an inducing point.} For the first problem, we define that, after discarding point \( u_{\mathrm{d}} \), the new joint distribution $\hat{q}(\bm{x}, \bm{f})$, still retains the factorized form \eqref{eq:approx_dist0}, namely, 

\begin{equation}
\begin{aligned}
    \hat{q}(\bm{x}, \bm{f}) = p(\bm{f}_{\ne u_{\mathrm{l}}}|\bm{u}_{\mathrm{l}}) \hat{q}(\bm{x}, \bm{u}_{\mathrm{l}})
\end{aligned}
\end{equation} 
where $p(\bm{f}_{\ne u_{\mathrm{l}}}|\bm{u}_{\mathrm{l}})$ is the conditional prior analogous to \eqref{eq:cond_prior}, and \( \hat{q}(\bm{x}, \bm{u}_{\mathrm{l}}) \) is defined as Gaussian to maintain closed-form learning capability.
Then, to ensure minimal accuracy loss, we can seek the optimal solution within the distribution family of \( \hat{q}(\bm{x}, \bm{f}) \), such that the distance between it and the original joint distribution \( q(\bm{x}, \bm{f}) = p(\bm{f}_{\ne u}|\bm{u})q(\bm{x}, \bm{u}) \) is minimized. Since \( \hat{q}(\bm{x}, \bm{u}_{\mathrm{l}}) \) controls the approximate distribution \( \hat{q}(\bm{x}, \bm{f}) \), it is sufficient to find the optimal distribution for \( \hat{q}(\bm{x}, \bm{u}_{\mathrm{l}}) \). Here, we use the inclusive KL divergence as a distance metric, namely \( \mathrm{KL}[q \| \hat{q}] = \int q \log \left( q/\hat{q} \right) \), which results in a simple optimal distribution:

\begin{equation}\label{eq:res_proj}
\begin{aligned}
& \hat{q}^*(\bm{x}, \bm{u}_{\mathrm{l}}) \\
&= \arg\min\limits_{\hat{q}(\bm{x}, \bm{u}_{\mathrm{l}})} 
\mathrm{KL} \left[ q(\bm{x}, \bm{f}) \| \hat{q}(\bm{x}, \bm{f}) \right] \\
&= \arg\min\limits_{\hat{q}(\bm{x}, \bm{u}_{\mathrm{l}})} 
\mathrm{KL} \left[ p(\bm{f}_{\ne u}|\bm{u})q(\bm{x}, \bm{u}) \| p(\bm{f}_{\ne u_{\mathrm{l}}}|\bm{u}_{\mathrm{l}}) \hat{q}(\bm{x}, \bm{u}_{\mathrm{l}}) \right] \\
&= \arg\min\limits_{\hat{q}(\bm{x}, \bm{u}_{\mathrm{l}})} 
\mathrm{KL} \left[q(\bm{x}, \bm{u}) \| p(u_{\mathrm{d}}|\bm{u}_{\mathrm{l}}) \hat{q}(\bm{x}, \bm{u}_{\mathrm{l}})\right] \\
&= \int q(\bm{x}, \bm{u}) \, \mathrm{d}u_{\mathrm{d}}
\end{aligned}
\end{equation}
It can be observed that the optimal distribution $\hat{q}^*(\bm{x}, \bm{u}_{\mathrm{l}})$ only involves marginalizing the discarded inducing point \( u_{\mathrm{d}} \) from the original distribution. Through \eqref{eq:res_proj}, optimal deletion for a given inducing point can be achieved.


\textbf{Criterion for selecting the least important inducing point.} 
Furthermore, for the second problem, the importance or novelty of a point can be assessed by the accuracy loss incurred from optimally discarding this point. Specifically, this is done by calculating the KL divergence in the third row of \eqref{eq:res_proj} with \( \hat{q}(\bm{x}, \bm{u}_{\mathrm{l}}) = \hat{q}^*(\bm{x}, \bm{u}_{\mathrm{l}}) \). The result is:

\begin{equation}\label{eq:Dstar}
\begin{aligned}
D^* &= \mathrm{KL} \left[q(\bm{x}, \bm{u}) \| p(u_{\mathrm{d}}|\bm{u}_{\mathrm{l}}) \hat{q}^*(\bm{x}, \bm{u}_{\mathrm{l}})\right] \\
& = D_1 - D_2 - D_3 \\
D_1 & = \int q(\bm{x}, \bm{u}) \log q(\bm{x}, \bm{u}) \, \mathrm{d}\bm{x} \, \mathrm{d}\bm{u} \\
D_2 &= \int q(\bm{u}) \log p(u_{\mathrm{d}} | \bm{u}_{\mathrm{l}}) \, \mathrm{d}\bm{u} \\
D_3 &= \int \hat{q}^*(\bm{x}, \bm{u}_{\mathrm{l}})  \log \hat{q}^*(\bm{x}, \bm{u}_{\mathrm{l}}) \, \mathrm{d}\bm{u}_{\mathrm{l}}
\end{aligned}
\end{equation}
where \( D^* \) quantifies the accuracy loss that results from discarding the inducing point \( u_{\mathrm{d}} \).

Based on \(D^*\), we can obtain a computable score to quantify the importance of inducing points. Specifically, we expand the integration in \eqref{eq:Dstar} and retain only the terms involving \(u_{\mathrm{d}}\), which yields the following score (for the derivation, see Appendix \ref{app:score}):


\begin{equation}\label{eq:score}
\begin{aligned}
s_d 
&= \Delta_1 + \Delta_2 + \Delta_3\\
\Delta_1 &= \bm{m}_u^T \bm{Q}_{du}^T \bm{Q}_{dd}^{-1} \bm{Q}_{du} \bm{m}_u \\
\Delta_2 &= \mathrm{tr}\left(\bm{Q}_{du} \bm{S}_{uu} \bm{Q}_{du}^T \bm{Q}_{dd}^{-1} \right) \\
\Delta_3 &= \log\vert \bm{\Omega}_{dd} \vert - \log\vert \bm{Q}_{dd} \vert
\end{aligned}
\end{equation}
where we denote the inverse of the joint covariance as \( \bm{\Omega} = \bm{\Sigma}_t^{-1} \), and \( \bm{\Omega}_{dd} \) is the element of the matrix \( \bm{\Omega} \) corresponding to the discarded point \( u_{\mathrm{d}} \), namely, its \((d_x + i_{\mathrm{d}})\)-th diagonal element. In addition, \( \bm{Q}_{dd} \) and \( \bm{Q}_{du} \) are the \( i_{\mathrm{d}} \)-th diagonal element and the \( i_{\mathrm{d}} \)-th row of the inverse kernel matrix \( \bm{Q} = \bm{K}_{uu}^{-1} \).
Among the terms in \eqref{eq:score}, \( \Delta_1 \) represents the accuracy loss in the mean of the joint distribution $q(\bm{x}_t, \bm{f})$, while \( \Delta_2 \) and \( \Delta_3 \) correspond to the loss in the covariance. 
As shown in \eqref{eq:score}, $\Delta_3$ involves the joint covariance of $(\bm x_t, \bm u)$, i.e., $\bm\Sigma_t$. Therefore, this score can take into account the correlation between the state $\bm x$ and the GP, and thus properly quantifies the accuracy loss in the context of GPSSMs.
Overall, the score \( s_d \) quantifies the relative accuracy loss from discarding \( u_{\mathrm{d}} \). A smaller score $s_d$ indicates a smaller accuracy loss. 
Therefore, it can be used as the criterion for selecting the least important point to discard.

Through \eqref{eq:res_proj} and \eqref{eq:score}, we have addressed the two main challenges for the selection of the inducing points.
Based on these, we can provide the adjustment rule for the inducing-point set, which is divided into discarding and adding operations.

\textbf{Rule for discarding inducing points.} For the discarding operation, the rule is as follows: if the size of the inducing-point set exceeds the budget \( M \), evaluate the score of each point using \eqref{eq:score} and remove the one with the lowest score. According to \eqref{eq:res_proj}, the moments of the new joint distribution \( \hat{q}^*(\bm{x}, \bm{u}_{\mathrm{l}}) \) can be easily evaluated by deleting the corresponding elements associated with \( u_{\mathrm{d}} \) from the original moments.

\textbf{Rule for adding inducing points.} 
For the adding operation, we use a more convenient and intuitive metric for selecting points, which is derived from the score in \eqref{eq:score}. Specifically, by observing the score \( s_d \) in \eqref{eq:score}, it can be seen that if \( \bm{Q}_{dd}^{-1} \to 0 \), the score will approach its minimum value, i.e., \( s_d \to -\infty \), indicating that the accuracy loss from deleting \( u_{\mathrm{d}} \) approaches 0. According to the matrix inversion formula (see Appendix A of \cite{csatoGaussianProcessesIterative2002}), \( \bm{Q}_{dd}^{-1} = \bm{K}_{dd} - \bm{K}_{dl} \bm{K}_{ll}^{-1} \bm{K}_{ld} \), where subscripts $d$ and $l$ correspond to $u_{\mathrm{d}}$ and $\bm u_{\mathrm{l}}$. In other words, $\bm Q_{dd}^{-1}$ is the GP prior conditional variance of \( u_{\mathrm{d}} \) given \( \bm{u}_{\mathrm{l}} \) according to \eqref{eq:GP_cond}. Therefore, we can use the GP prior conditional variance of \( f_t \) given the current inducing points $\bm u$ as the metric for novelty, namely:

\begin{equation}\label{eq:gam}
\begin{aligned}
\gamma = \bm{K}_{tt} - \bm{K}_{tu} \bm{K}_{uu}^{-1} \bm{K}_{ut}
\end{aligned}
\end{equation}
where subscript $t$ denotes $f_t$.
Then, if \( \gamma \) is less than a certain threshold, denoted as \( \varepsilon_{\text{tol}} \), we will not add the new point to the inducing-point set. 
Through this adding rule, we obtain two advantages. First, this filtering process helps to slow the increase in the size of the inducing-point set. Second, it plays a crucial role in ensuring the numerical stability of the learning algorithm.
Specifically, when \( \gamma \) is small, adding a new point can cause the updated kernel matrix \( \bm{K}_{\bar{u} \bar{u}} \) to approach singularity, which would negatively affect the stability of the algorithm. Therefore, the value of the threshold $\varepsilon_{\text{tol}}$ for adding a point can be determined based on the machine accuracy.

In the above adding rule, there exist cases where points are not added. In this situation, the moment updating equation \eqref{eq:pred} is no longer applicable. Therefore, we re-derive the moment updating equation for this scenario. The derivation details are provided in Appendix \ref{app:pred} and the result is as follows:

\begin{equation}\label{eq:pred1}
\begin{aligned}
& \bm{\xi}_{t+1}^- = [\bm{F}_t^T, \bm{m}_{\bm{u}}^T]^T \\
& \bm{\Sigma}_{t+1}^- = \bm{\Phi} \bm{\Sigma}_t \bm{\Phi}^T + \bm{\Sigma}_{p,\bm{X}} \\
& \bm{\Phi} = \begin{bmatrix}
\bm{A}_x & \bm{A}_f \bm{k}_{tu} \\
\bm{0} & \bm{I}_{n_u} \\
\end{bmatrix} \\
& \bm{\Sigma}_{p,\bm{X}} = \begin{bmatrix}
\bm{A}_f \gamma \bm{A}_f^T + \bm{\Sigma}_p & \bm{0} \\
\bm{0} & \bm{0} \\
\end{bmatrix}
\end{aligned}
\end{equation}
This equation is similar to the original equation \eqref{eq:pred}, differing only in the expression for the transition Jacobian \( \bm{\Phi} \) and the process noise covariance \( \bm{\Sigma}_{p,\bm{X}} \).

In summary, this subsection derives the optimal deletion method for inducing points and presents a metric for quantifying the importance or novelty of points. Based on these, a dynamic adjustment algorithm, including discarding and adding operations for the inducing-point set, is developed to ensure limited computational cost and algorithm stability. In the next subsection, we will address the online hyperparameter optimization problem to enhance learning accuracy.

\subsection{Online Optimization of GP Hyperparameters}\label{subsec:opt_hp}

In GP learning, if the hyperparameters \( \bm{\theta} \) significantly mismatch the function to be learned, it will result in substantial learning errors. In offline learning, the hyperparameters \( \bm{\theta} \) can be optimized using data by maximizing the log marginal likelihood \cite{rasmussenGaussianProcessesMachine2008}. 
However, as illustrated in the Introduction, the online setting presents two challenges for hyperparameter optimization: 1) the information source problem due to the lack of data retention, and 2) the coupling between GP hyperparameters and the posterior distribution.
In this subsection, we will address these two challenges and propose an online GP hyperparameter optimization method for GPSSMs. For clarity, we denote the quantities corresponding to before and after hyperparameter optimization with the superscripts or subscripts "old" and "new", respectively. Additionally, to highlight the impact of the GP hyperparameters, probability distributions depending on \( \bm{\theta} \) will be rewritten as \( p(\cdot; \bm{\theta}) \).

\textbf{Information source for online hyperparameters optimization.}
Firstly, to find the information source for hyperparameter optimization, we can first observe the expression of the exact joint posterior distribution, which is given as follows for time $t$:

\begin{equation}\label{eq:joint_t} 
\begin{aligned} 
p(\bm{x}_{t}, \bm{f}|\bm{y}_{1:t}; \bm{\theta})
&= \dfrac{p(\bm{f}; \bm{\theta}) p(\bm{x}_t, \bm{y}_{1:t}|\bm{f}) }
{p(\bm{y}_{1:t}; \bm{\theta})}
\end{aligned} 
\end{equation}
where \( p(\bm{f}; \bm{\theta}) \) denotes the GP prior on function values \( \bm{f} \), and the likelihood model \( p(\bm{x}_t, \bm{y}_{1:t}|\bm{f}) = 
\int \prod_{\tau=1}^{t} p(\bm{y}_\tau, \bm{x}_\tau|\bm{x}_{\tau-1}, \bm{f}) p(\bm{x}_0) \, \mathrm{d} \bm{x}_{0:t-1}\). 
By using this likelihood model, we can evaluate the marginal likelihood \( p(\bm{y}_{1:t}; \bm{\theta}) = \int p(\bm{f}; \bm{\theta}) p(\bm{x}_t, \bm{y}_{1:t}|\bm{f}) \, \mathrm{d} \bm{x}_t \, \mathrm{d} \bm{f} \). Then, the GP hyperparameters $\bm \theta$ can be optimized by maximizing the marginal likelihood $p(\bm{y}_{1:t}, \bm{\theta})$.
However, for online hyperparameter optimization, an explicit likelihood model is unavailable because the past measurements \( \bm{y}_{1:t-1} \) are not stored. Instead, we only have an approximation of the left-hand side of \eqref{eq:joint_t}, namely the approximate distribution corresponding to the old hyperparameters $\bm\theta_{\mathrm{old}}$:

\begin{equation}
\begin{aligned}
    q_{\text{old}}(\bm{x}_t, \bm{f}) \approx p(\bm{x}_{t}, \bm{f}|\bm{y}_{1:t}; \bm{\theta}_{\text{old}}) 
\end{aligned}
\end{equation}
Fortunately, by observing \eqref{eq:joint_t}, it can be found that the likelihood model can be implicitly obtained by:

\begin{equation}\label{eq:like}
\begin{aligned}
p(\bm{x}_t, \bm{y}_{1:t}|\bm{f}) &= \dfrac{p(\bm{x}_{t}, \bm{f}|\bm{y}_{1:t}; \bm{\theta}_{\text{old}}) p(\bm{y}_{1:t};\bm{\theta}_{\text{old}}) }
{p(\bm{f};\bm{\theta}_{\text{old}})} \\
&\approx \dfrac{q_{\text{old}}(\bm{x}_t, \bm{f}) p(\bm{y}_{1:t};\bm{\theta}_{\text{old}}) }
{p(\bm{f};\bm{\theta}_{\text{old}})} \\
\end{aligned}
\end{equation}
In other words, the measurement information is distilled into the approximate posterior \(q_{\mathrm{old}}(\bm{x}_t, \bm{f})\) and the latter can recover the likelihood model, thereby addressing the information source problem. 

\textbf{Posterior update equation.} 
Secondly, by utilizing the likelihood model obtained in \eqref{eq:like}, we can find the optimization objective for GP hyperparameters and evaluate the posterior distribution after hyperparameter update.
The key idea in achieving this is to evaluate the joint probability model $p(\bm{f}; \bm{\theta}) p(\bm{x}_t, \bm{y}_{1:t}|\bm{f})$, which is proportional to the posterior distribution, and its integration is the marginal likelihood that can serve as the objective function. For convenience, we derive the posterior update equation first, which can be obtained by combining the expression of the posterior distribution \eqref{eq:joint_t} with the approximate likelihood model \eqref{eq:like}, namely:


\begin{equation}\label{eq:qxf_hp}
\begin{aligned}
& p(\bm{x}_{t}, \bm{f}|\bm{y}_{1:t}; \bm{\theta}_{\text{new}}) \\
&= \dfrac{p(\bm{f}; \bm{\theta}_{\text{new}}) p(\bm{x}_t, \bm{y}_{1:t}|\bm{f})}
{p(\bm{y}_{1:t}; \bm{\theta}_{\text{new}})} \\
&\approx \dfrac{p(\bm{f}; \bm{\theta}_{\text{new}}) 
q_{\text{old}}(\bm{x}_t, \bm{f}) p(\bm{y}_{1:t};\bm{\theta}_{\text{old}})}
{p(\bm{y}_{1:t}; \bm{\theta}_{\text{new}}) p(\bm{f};\bm{\theta}_{\text{old}})} \\
&\triangleq q_{\text{new}}(\bm{x}_t, \bm{f})
\end{aligned}
\end{equation}
where \( q_{\text{new}}(\bm{x}_t, \bm{f}) \) denotes the approximate joint distribution corresponding to the new hyperparameters \( \bm{\theta}_{\text{new}} \). In \eqref{eq:qxf_hp}, since the distributions \( p(\bm{f}; \bm{\theta}_{\text{old}}) \), \( p(\bm{f}; \bm{\theta}_{\text{new}}) \) and \( q_{\text{old}}(\bm{x}_t, \bm{f}) \) are all Gaussian, the new approximate distribution \( q_{\text{new}}(\bm{x}_t, \bm{f}) \) is also Gaussian. By expanding the distribution related to \( \bm{f} \) in the third row of \eqref{eq:qxf_hp} with the conditional distribution \( p(\bm{f}_{\ne u}|\bm{u}) \), it can be shown that the new distribution retains the factorized form \( q_{\text{new}}(\bm{x}_t, \bm{f}) = p(\bm{f}_{\ne u}|\bm{u}; \bm{\theta}_{\text{new}}) q_{\text{new}}(\bm{x}_t, \bm{u}) \), where:

\begin{equation}\label{eq:qnew}
\begin{aligned}
q_{\text{new}}(\bm{x}_t, \bm{u}) &= \dfrac{p(\bm{u}; \bm{\theta}_{\text{new}}) 
q_{\text{old}}(\bm{x}_t, \bm{u}) p(\bm{y}_{1:t};\bm{\theta}_{\text{old}})}
{p(\bm{y}_{1:t}; \bm{\theta}_{\text{new}}) p(\bm{u};\bm{\theta}_{\text{old}})} \\
&\propto 
\dfrac{p(\bm{u}; \bm{\theta}_{\text{new}}) q_{\text{old}}(\bm{x}_t, \bm{u})}
{p(\bm{u};\bm{\theta}_{\text{old}})}
\end{aligned}
\end{equation}
Therefore, we can obtain the new posterior distribution after the GP hyperparameters change. In practical implementation, this can be achieved by a Kalman filter-like equation (the derivation is provided in Appendix \ref{app:hp}):

\begin{equation}\label{eq:update_hp}
\begin{aligned}
&\tilde{\bm{H}} =  \begin{bmatrix} \bm{0} & \bm{I}_{n_u} \end{bmatrix} \\
&\tilde{\bm{G}} = \bm{\Sigma}_{t}^{\text{old}} \tilde{\bm{H}}^T \left( \bm{S}_{uu}^{\text{old}} + \Delta \bm{K} \right)^{-1} \\
&\bm{\xi}_{t}^{\text{new}} = \bm{\xi}_{t}^{\text{old}} - \tilde{\bm{G}} \bm{\xi}_{t}^{\text{old}} \\
&\bm{\Sigma}_{t}^{\text{new}} = \bm{\Sigma}_{t}^{\text{old}} - \tilde{\bm{G}} \tilde{\bm{H}} \bm{\Sigma}_{t}^{\text{old}}
\end{aligned}
\end{equation}

\textbf{Optimization objective.}
Furthermore, by integrating both sides of the first row of \eqref{eq:qnew}, we can obtain the marginal likelihood for hyperparameter optimization, which leads to:

\begin{equation}\label{eq:py_new}
\begin{aligned}
\dfrac{p(\bm{y}_{1:t};\bm{\theta}_{\text{new}})} 
{p(\bm{y}_{1:t}; \bm{\theta}_{\text{old}})} 
&= \int \dfrac{p(\bm{u}; \bm{\theta}_{\text{new}}) q_{\text{old}}(\bm{x}_t, \bm{u})}
{p(\bm{u};\bm{\theta}_{\text{old}})} \, \mathrm{d} \bm{x}_t \, \mathrm{d} \bm{u}
\end{aligned}
\end{equation}
Then, by expanding the probability density in \eqref{eq:py_new} and eliminating terms that are irrelevant to \( \bm{\theta}_{\text{new}} \) (the derivation is provided in Appendix \ref{app:hp}), a practical optimization objective for GP hyperparameters is to minimize:



\begin{equation}\label{eq:loss}
\begin{aligned}
&\mathcal{L} = \mathcal{L}_1 + \mathcal{L}_2 \\
&\mathcal{L}_1 = (\bm{m}_{u}^{\text{old}})^T (\bm{S}_{uu}^{\text{old}} + \Delta \bm{K})^{-1} \bm{m}_{u}^{\text{old}} \\
&\mathcal{L}_2 = \log \vert \bm{K}_{uu}^{\text{new}} + [ \bm{I}_{n_u} - \bm{K}_{uu}^{\text{new}} (\bm{K}_{uu}^{\text{old}})^{-1} ] \bm{S}_{uu}^{\text{old}} \vert \\
&\Delta \bm{K} = [(\bm{K}_{uu}^{\text{new}})^{-1} - (\bm{K}_{uu}^{\text{old}})^{-1} ]^{-1} \\
\end{aligned}
\end{equation}
where \( \bm{K}_{uu}^{\text{new}} \) and \( \bm{K}_{uu}^{\text{old}} \) denote the kernel matrix of \( \bm{u} \) evaluated with new and old GP hyperparameters. Therefore, the GP hyperparameters can be optimized online by implementing gradient descent on the loss function \eqref{eq:loss} in each algorithm iteration. 
Since the approximate posterior \( q(\bm{x}, \bm{f}) \) retains the measurement information, it is not necessary to optimize to convergence in a single algorithm iteration.

In summary, by recovering an approximate likelihood model from the current filtering distribution, this subsection derives the  objective function and posterior update equation for online hyperparameter optimization. This method reduces the burden of hyperparameter tuning at the learning initialization stage and improves online learning accuracy by adjusting the GP prior to match the actual function characteristics.

\begin{remark}
The online optimization method for GP hyperparameters most closely related to this paper is \cite{buiStreamingSparseGaussian2017}, an online GPR method. In this method, the likelihood model is similarly recovered from the approximate posterior distribution, and it is used to optimize the location of inducing points and GP hyperparameters. Clearly, in the context of GPSSMs, it is possible to optimize the inducing inputs in the same way. 
However, we do not pursue this approach because simultaneously optimizing the inducing inputs and hyperparameters would complicate the loss function and the posterior update equation. Additionally, we observe that in GPSSMs, due to the system states being implicitly measured, the difficulty of optimizing the inducing inputs increases. Therefore, we separate the optimization of inducing inputs and GP hyperparameters and determine the inducing-point set using a discretized selection method.
\end{remark}

\subsection{Algorithm Summary}

\begin{figure*}[h]
    \centering
    \includegraphics[width=0.96\textwidth]{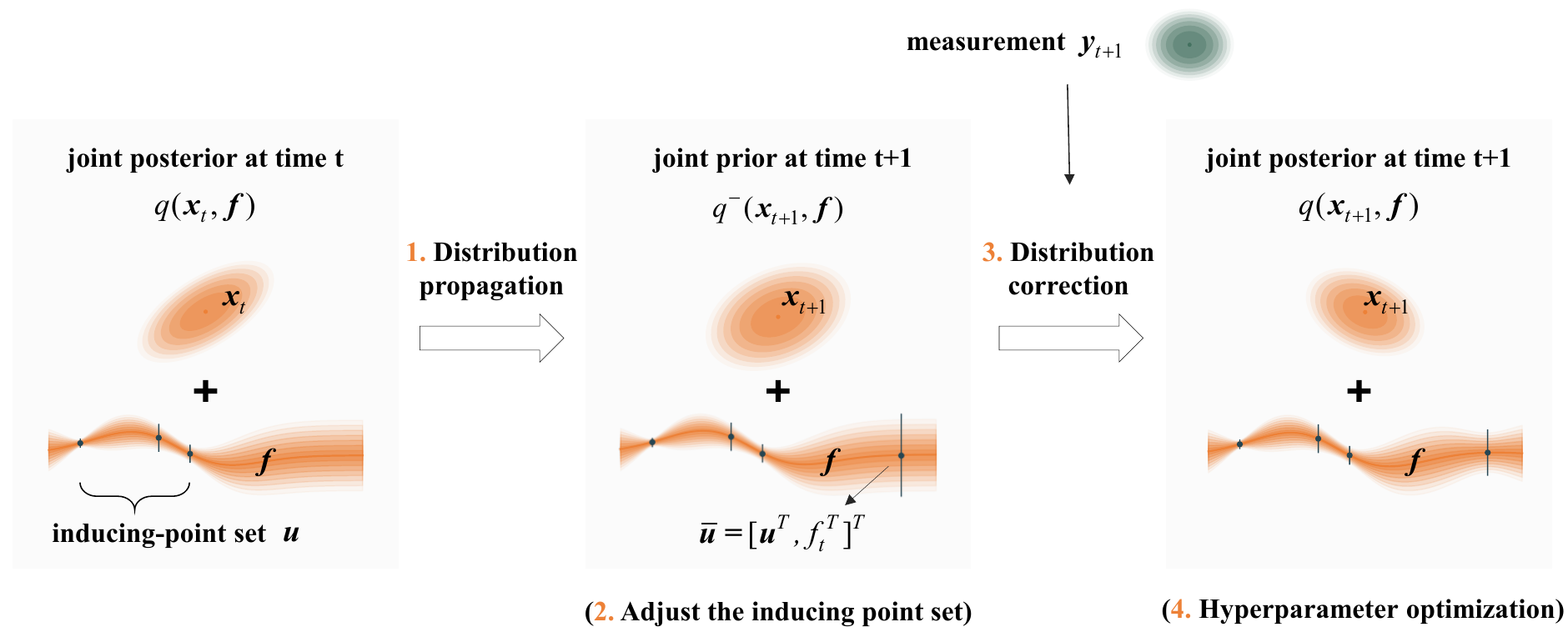}
    \caption{Schematic diagram of the RGPSSM algorithmic workflow.}
    \label{fig:alg_flow}
\end{figure*}

In summary, we have addressed the four challenges outlined in the Introduction for implementing online GPSSMs. Specifically, first, nonlinearity and coupling are handled through first-order linearization and joint inference of the system state and GP, leading to a closed-form Bayesian update equation. Second, a dynamic inducing-point adjustment algorithm is developed to overcome the computational burden caused by the non-parametric nature of the GP. Third, we recover the likelihood model from the approximate posterior distribution and utilize it for the online optimization of GP hyperparameters, thus enhancing learning accuracy. Based on these solutions, we have developed an efficient online GPSSM method that can adapt to both the operating domain and GP hyperparameters. 
Compared to existing online GPSSM methods, the proposed method is the first to achieve recursive inference for GPSSMs with the ability to adapt to both the operating domain and GP hyperparameters. This makes it truly applicable to various flexible learning scenarios.
Therefore, we refer to this method as the recursive GPSSM (RGPSSM).
The pseudocode of RGPSSM is summarized in Algorithm \ref{alg:rgpssm}, and the schematic diagram of the algorithm workflow is shown in Fig.~\ref{fig:alg_flow}.


\begin{algorithm}[h]
    \caption{Recursive Gaussian Process State Space Model (RGPSSM)}
    \label{alg:rgpssm}
    \begin{footnotesize}
    1: \textbf{Input:} initial inducing points $\bm{u}$, threshold for adding points $\varepsilon_{tol}$, budget for the inducing-point set $M$.\\
    2: \textbf{Repeat until operation ends:} \\
    3: \quad Assess the novelty $\gamma$ of the new point $f_t$ using \eqref{eq:gam}.\\
    4: \quad \textbf{If} $\gamma > \varepsilon_{tol}$ \textbf{then} \\
    5: \qquad \parbox[t]{0.8\linewidth}{Add $f_t$ to the inducing-point set and propagate the moments using \eqref{eq:pred}.}\\
    6: \quad \textbf{Else} \\
    7: \qquad \parbox[t]{0.8\linewidth}{Propagate the moments using \eqref{eq:pred1}.}\\
    8: \quad \textbf{If} the number of inducing points $n_u > M$ \textbf{then} \\
    9: \qquad \parbox[t]{0.8\linewidth}{Identify the least important point based on the score in \eqref{eq:score} and remove it using \eqref{eq:res_proj}.}\\
    10: \quad \textbf{If} a new measurement $y_t$ is available \textbf{then} \\
    11: \qquad \parbox[t]{0.8\linewidth}{Correct the moments using \eqref{eq:corr}.}\\
    12: \quad \parbox[t]{0.8\linewidth}{Minimize $\mathcal{L}$ in \eqref{eq:loss} to update the GP hyperparameters $\bm{\theta}$ using the Adam optimizer \cite{kingma2014adam}, and subsequently update the moments using \eqref{eq:update_hp}.}\\
    \end{footnotesize}
\end{algorithm}

In theory, Algorithm \ref{alg:rgpssm} has recursive learning capability for GPSSMs. However, when implemented on a computer, there exists numerical stability issues as in the Kalman filter.
Specifically, the algorithm requires the evolution of the covariance matrix \( \bm{\Sigma}_t \) of the augmented state \( \bm{X}_t \), which can easily lose positive definiteness due to the accumulation of rounding errors. 
Compared to general KF applications, this issue is relatively severe in the context of RGPSSM because the dimension of the joint covariance \( \bm{\Sigma}_t \) is often higher.
To address this problem, a stable implementation based on Cholesky factorization is derived, as detailed in Appendix \ref{app:stable}. This implementation method is capable of enhancing the numerical stability by making the computation more compact. In the next section, we will discuss the computational complexity of the Cholesky version of RGPSSM, along with the extension to the multi-output GP cases.

\section{Extension to Multi-output GP and Discussion}

This section extends the single-output GP version of the RGPSSM method to the multi-output GP case, and systematically analyzes the computational complexity, generalization, and potential of the proposed method.

\subsection{Multi-output GP Version}\label{subsec:multi_output}

In numerous practical applications, the unknown function to be learned within SSMs exhibits multiple outputs, represented as $\bm{f(\cdot)}: \mathbb R^{d_x} \to \mathbb R^{d_f}$ where $d_f > 1$. To accommodate this multi-output nature, the GP modeling for $\bm{f(\cdot)}$ must be extended to a multi-output form, necessitating a modification of the kernel function $\bm{K}_0(\cdot, \cdot)$ to yield a $d_f \times d_f$ matrix output. For this purpose, we employ a straightforward multi-output kernel formulation, expressed as:

\begin{equation}\label{eq:multi_output_kernel}
\begin{aligned}
    \bm{K}_0(\cdot, \cdot) = {K}_0(\cdot, \cdot)\, \bm{\Xi}
\end{aligned}
\end{equation}
where ${K}_0(\cdot, \cdot)$ in the RHS represents the single-output kernel, and $\bm{\Xi}$ denotes the signal covariance between each function output dimension. 
According to the multi-output GP literature \cite{alvarez2012kernels}, this kernel belongs to the linear model of coregionalization (LMC) method.
Based on this multi-output kernel function, the kernel matrix has a Kronecker product structure, i.e., $\bm{K}_0(\mathcal{X}_1, \mathcal{X}_2) = {K}_0(\mathcal{X}_1, \mathcal{X}_2) \otimes \bm{\Xi}$. This structure indicates that the $(i,j)$-th $d_f \times d_f$ block of the kernel matrix $\bm{K}_0(\mathcal{X}_1, \mathcal{X}_2)$ is the prior covariance between function values $\bm{f}(\textbf{x}_i)$ and $\bm{f}(\textbf{x}_j)$. Therefore, the only modification from the single-output GP version of RGPSSM to the multi-output GP version is to convert scalar elements in related vectors or matrices into corresponding blocks, which are presented in the following.

Firstly, the moments of joint distribution $q(\bm x, \bm u)$ including mean $\bm m_{u}$, cross-covariance $\bm V_{x u}$ and auto-covariance $\bm S_{uu}$ have new definitions. Specifically,
\begin{itemize}
    \item The $i$-th column block (with shape $d_f \times 1$) of column vector $\bm m_{u}$ represents the mean of the inducing point $\bm u_i$
    \item The $i$-th row block (with shape $d_x \times d_f$) of matrix $\bm V_{x u}$ represents the cross-covariance between state $\bm x$ and the inducing point $\bm u_i$
    \item The $(i, j)$-th block (with shape $d_f \times d_f$) of matrix $\bm S_{uu}$ represents the cross-covariance between the inducing points $\bm u_i$ and $\bm u_j$
\end{itemize}

Secondly, among the inducing point management, the calculation of the score $s_d$ and novelty $\gamma$ require updates. For the evaluation of the score $s_d$ in \eqref{eq:score}, the modifications only involve updating the definitions of several quantities: 
\begin{itemize}
    \item $\bm Q_{du}$ is the $i_{\mathrm{d}}$-th row block of $\bm Q$
    \item $\bm Q_{dd}$ is the $i_{\mathrm{d}}$-th diagonal block of $\bm Q$
    \item $\bm\Omega_{dd}$ is the $i_{\mathrm{d}}$-th diagonal block of the inducing-point submatrix in $\Omega$, where the inducing-point submatrix is defined as the matrix slice $\bm\Omega[d_x:, d_x:]$
\end{itemize} 
For the criterion of adding points, since $\bm\gamma$ in \eqref{eq:gam} is now a matrix, the criterion is modified to $\mathrm{tr}(\bm\gamma) > \varepsilon_{tol}$. Additionally, when adding or removing inducing points, the updates to the moments are performed in a block-wise manner based on the new definitions.

Therefore, the extension to the multi-output GP version is straightforward, with modifications involving the kernel function, the definitions of some vectors and matrices, and the evaluation criteria for inducing points management. 
It is evident that the dimensions of the statistical moments scale with the output dimension of the GP, thus impacting the computational complexity. This is analyzed in the next subsection.

\subsection{Computational Complexity Analysis}

The computational complexity of RGPSSM is analyzed for its various algorithmic steps in Table \ref{tab:complexity}, which presents a detailed comparison between the standard and Cholesky-based implementations of RGPSSM. This analysis is based on the multi-output GP version of RGPSSM.

\begin{table*}[h]
\centering
\caption{Computational complexity comparison between standard and Cholesky versions of RGPSSM}
\begin{tabular}{lcccc}
\hline
Algorithm step & Standard RGPSSM & Involved equations & Cholesky RGPSSM & Involved equations \\
\hline
GP prediction & $\mathcal{O}\left( (d_f M)^2 \right)$ & \eqref{eq:qf} 
& $\mathcal{O}\left( (d_f M)^2 \right)$ & \eqref{eq:qf} \\
Moment propagation & $\mathcal{O}\left( (d_x + d_f M)^2 \right)$ & \eqref{eq:pred}, \eqref{eq:pred1} 
& $\mathcal{O}\left( (d_x + d_f M)^3 \right)$ & \eqref{eq:qr0}-\eqref{eq:chol_pred2} \\
Inducing point removal & $\mathcal{O}\left( (d_x + d_f M)^3 \right)$ & \eqref{eq:score} 
& $\mathcal{O}\left( (d_x + d_f M)^3 \right)$ & \eqref{eq:score}, \eqref{eq:chol_discard} \\
Moment correction & $\mathcal{O}\left( (d_x + d_f M)^2 \right)$ & \eqref{eq:corr} 
& $\mathcal{O}\left( (d_x + d_f M)^2 \right)$ & \eqref{eq:chol_post} \\
Hyperparameter update & $\mathcal{O}\left( (d_x + d_f M)^3 \right)$ & \eqref{eq:loss}, \eqref{eq:update_hp} 
& $\mathcal{O}\left( (d_x + d_f M)^3 \right)$ & \eqref{eq:loss}, \eqref{eq:update_hp}, \eqref{eq:chol_hp} \\
\hline
Total complexity & $\mathcal{O}\left( (d_x + d_f M)^3 \right)$ & - 
& $\mathcal{O}\left( (d_x + d_f M)^3 \right)$ & - \\
\hline
\end{tabular}
\label{tab:complexity}
\end{table*}

Based on the complexity analysis in Table \ref{tab:complexity}, the following additional remarks are provided:

\begin{itemize}
    \item By precomputing and storing the Cholesky factor of the kernel matrix $\bm K_{uu}$ during updates, we can avoid repeated matrix inversions in GP prediction \eqref{eq:qf}, thereby reducing the computational complexity of the GP prediction to quadratic order.
    \item As shown in Table \ref{tab:complexity}, both the standard and Cholesky versions of RGPSSM exhibit cubic total complexity. Although the Cholesky version incurs higher computational costs in certain algorithmic steps, it offers superior numerical stability. Therefore, we recommend the Cholesky version for general applications, which is adopted throughout all experimental evaluations.
    \item It is clear that the overall algorithmic complexity grows cubically with both the function output dimension $d_f$ and the number of inducing points $M$. 
    This is mainly due to two factors: the EKF-like moment matching, which provides second-order learning speed \cite{eon1998online}, and the flexible inducing point adjustment and hyperparameter optimization algorithms.
     Although this cubic complexity limits the algorithm's applicability to high-dimensional scenarios, it offers a stable, fast, and flexible learning solution for low-dimensional problems, which are common in many applications such as aerospace \cite{singh1995direct} and robotics \cite{oconnellNeuralFlyEnablesRapid2022}.
\end{itemize}

\subsection{Generalization and Applicability}

RGPSSM can be regarded as a unified generalization of several existing methods. Specifically, it subsumes both the pre-parameterized AKF-based online GPSSM with inducing points approximation \cite{Kullberg2020} and kernel-based online regression approaches such as KRLS \cite{VanVaerenbergh2012} and SOGP \cite{csatoGaussianProcessesIterative2002} as special cases. For example, if RGPSSM is initialized with a fixed set of inducing points and no further updates are performed (i.e., by setting \( \varepsilon_{\text{tol}} = +\infty \)), it reduces to the pre-parameterized AKF-based method. Furthermore, when the transition and measurement functions in the GPSSM are set as \( x_{t} = f(\bm{c}_{t}) \) and \( y_t = x_t \), where \( \bm{c}_t \) and \( y_t \) denote the function input and output, respectively, RGPSSM becomes equivalent to KRLS and SOGP. Therefore, RGPSSM offers practical and flexible recursive learning capabilities applicable to a wide range of scenarios, as summarized below:

\begin{itemize}
    \item Applicable to various observation scenarios, including partial state observation ($\bm y_t = \bm{g}(\bm{x}_t)$), full state observation ($\bm y_t = \bm{x}_t$), and direct function value observation ($\bm y_t = \bm{f}(\bm{x}_t)$), as well as any combination of these cases.
    \item Benefiting from the two-step inference formulation, it is suitable for cases involving irregular measurement intervals or measurement losses.
    \item Since it belongs to the AKF framework, it can be seamlessly integrated with existing adaptive Kalman filtering methods to enable adaptation of process and measurement noise covariances.
    \item The use of first-order linearization, together with the augmentation of the inducing-point set, enables a finite-dimensional Gaussian approximation of the joint state–inducing point distribution. This provides a flexible foundation for incorporating advanced nonlinear filters, such as the sigma point filter \cite{wan2000unscented,arasaratnam2009cubature}, iterative filter \cite{kullberg2024dynamically}, variational filter \cite{dowling2024exponential}, or exact moment evaluation \cite{deisenroth2009analytic}, to further improve the accuracy of moment matching.
\end{itemize}

In the next section, we will validate the effectiveness and advantages of RGPSSM through several experiments.

\section{Experiment Result}

This section demonstrates the proposed method using two synthetic simulations and a real-world dataset. Section \ref{subsec:exp1} presents a comparative experiment on a synthetic NASCAR$^\circledR$ dataset, which serves as a benchmark for online GPSSM methods. Section \ref{subsec:exp2} showcases the capability of RGPSSM for hyperparameter and inducing point adaptation in a synthetic wing rock dynamics learning task. Section \ref{subsec:exp3} evaluates the learning performance of RGPSSM on real-world system identification tasks.

The learning algorithm is implemented on a desktop computer running Python~3.9, 
with an Intel(R) Core(TM) i7-14700F 2.10~GHz processor and 32~GB of RAM, 
and the associated code is published online\footnote{\url{https://github.com/TengjieZheng/rgpssm}}.
In all experiments, the multi-output kernel \eqref{eq:multi_output_kernel} is employed, where the single-output kernel is the commonly used Squared Exponential Automatic Relevance Determination (SEARD) \cite{rasmussenGaussianProcessesMachine2008} kernel, which is defined as:

\begin{equation}
    \begin{aligned}
    K_0(\mathbf{x}, \mathbf{x}^\prime) = \exp\left(-\dfrac{1}{2}
    (\mathbf{x} - \mathbf{x}^\prime)^T \bm{\Lambda}^{-1} (\mathbf{x} - \mathbf{x}^\prime)
     \right)
    \end{aligned}
\end{equation}
where the diagonal matrix \( \bm{\Lambda} = \mathrm{diag}(l_1^2, l_2^2, \dots, l_{d_x}^2) \) encodes the length scales for each input dimension. The signal covariance matrix \(\bm{\Xi}\) is defined as $\mathrm{diag}(\sigma_1^2, \sigma_2^2, \dots, \sigma_{d_f}^2)$, where each $\sigma_i^2$ denotes the signal variance corresponding to the $i$-th output dimension of the function. This kernel formulation enables the model to flexibly characterize both the input-output relevance of different features (via $l_i$) and the signal strength of each output (via $\sigma_i^2$). All these hyperparameters will be optimized online to adaptively capture the underlying function characteristics during learning.

\subsection{Synthetic NASCAR$^\circledR$ Dynamics Learning}\label{subsec:exp1}

To benchmark the online learning performance of RGPSSM, we compare it with two state-of-the-art (SOTA) online GPSSM algorithms: SVMC \cite{zhaoStreamingVariationalMonte2023}, which represents the PF-based methods, and OEnVI \cite{Lin2024}, which exemplifies the SVI-based methods.
 Since the pre-parameterized AKF-based method is a special case of the proposed method, it is not included in the comparison.  
In the experiment, all methods are tested using synthetic NASCAR\textsuperscript{\textregistered} data \cite{lindermanBayesianLearningInference2017}. This dataset is a benchmark for online GPSSM methods and is used in \cite{zhaoStreamingVariationalMonte2023} and \cite{Lin2024}.
This dataset involves a two-dimensional state with dynamics that follow a recurrent switching linear dynamical system (see \cite{lindermanBayesianLearningInference2017} for more details).
The transition model is $\bm x_{t+1} = \bm x_t + \bm f(\bm x_t)$, where the function $\bm f(\bm x_t)$ is defined by the dataset in \cite{lindermanBayesianLearningInference2017} and is learned by the GP. The measurement model is given by \( \bm{y}_t = \bm{C}\bm{x}_t + \bm{w}_t \), where \( \bm{w}_t \sim N(\bm{0}, 0.1^2 \bm{I}_4) \), $d_y=4$, and \( \bm{C} \) is a 4-by-2 matrix with random elements.
The three methods are trained with 500 measurements and tested by predicting the subsequent 500 steps. Note that, unlike the experiments in \cite{zhaoStreamingVariationalMonte2023} and \cite{Lin2024}, we use fewer dimensions and a smaller number of measurements for training, providing a more rigorous evaluation. 
For a fair comparison, the maximum number of inducing points across all methods is limited to 20, and the GP hyperparameters are set to the same values and are not adjusted online. The implementation and other parameter settings for the SVMC and OEnVI algorithms 
are based on the code provided online\footnote{\url{https://github.com/catniplab/svmc}}%
\footnote{\url{https://github.com/zhidilin/gpssmProj}}.

\begin{figure*}[h]
    \centering
    \subfloat[RGPSSM]{%
        \includegraphics[width=0.3\textwidth]{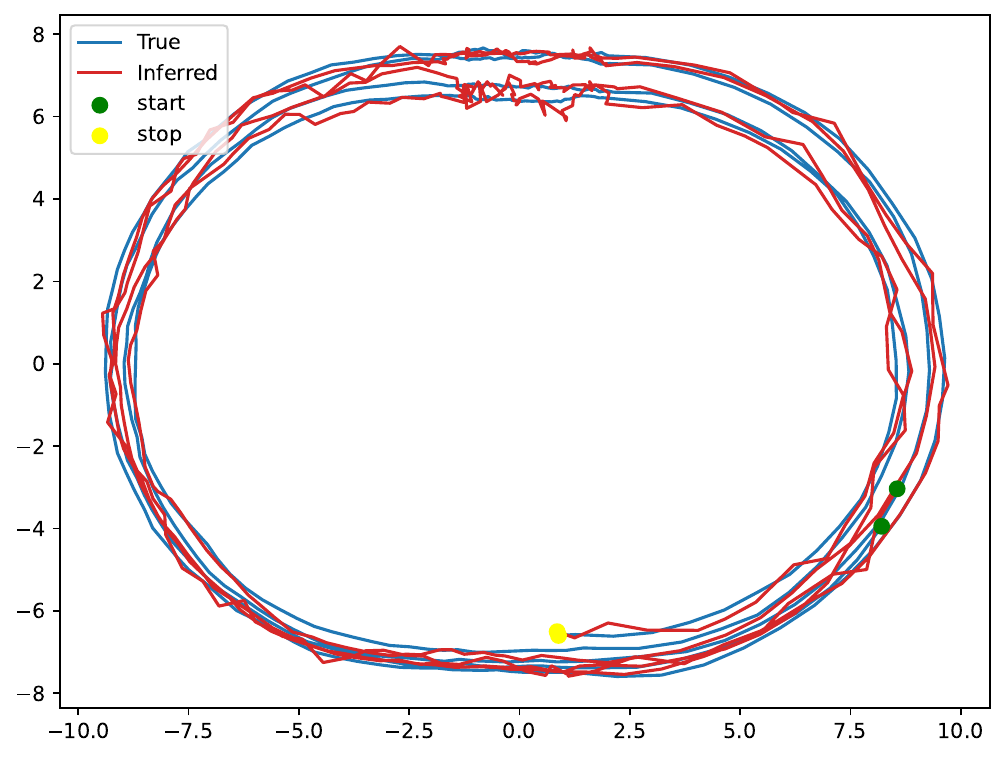}
        \label{fig:nascar_sub1}
    }
    \hspace{0.005\textwidth} 
    \subfloat[SVMC]{%
        \includegraphics[width=0.3\textwidth]{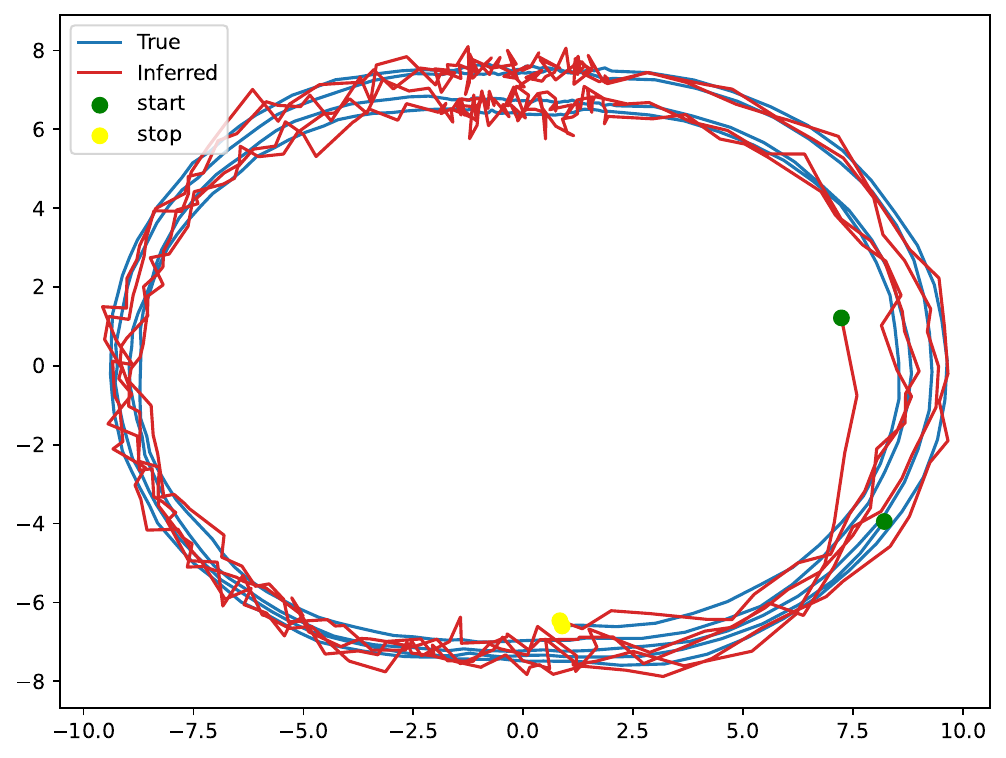}
        \label{fig:nascar_sub2}
    }
    \hspace{0.005\textwidth}
    \subfloat[OEnVI]{%
        \includegraphics[width=0.3\textwidth]{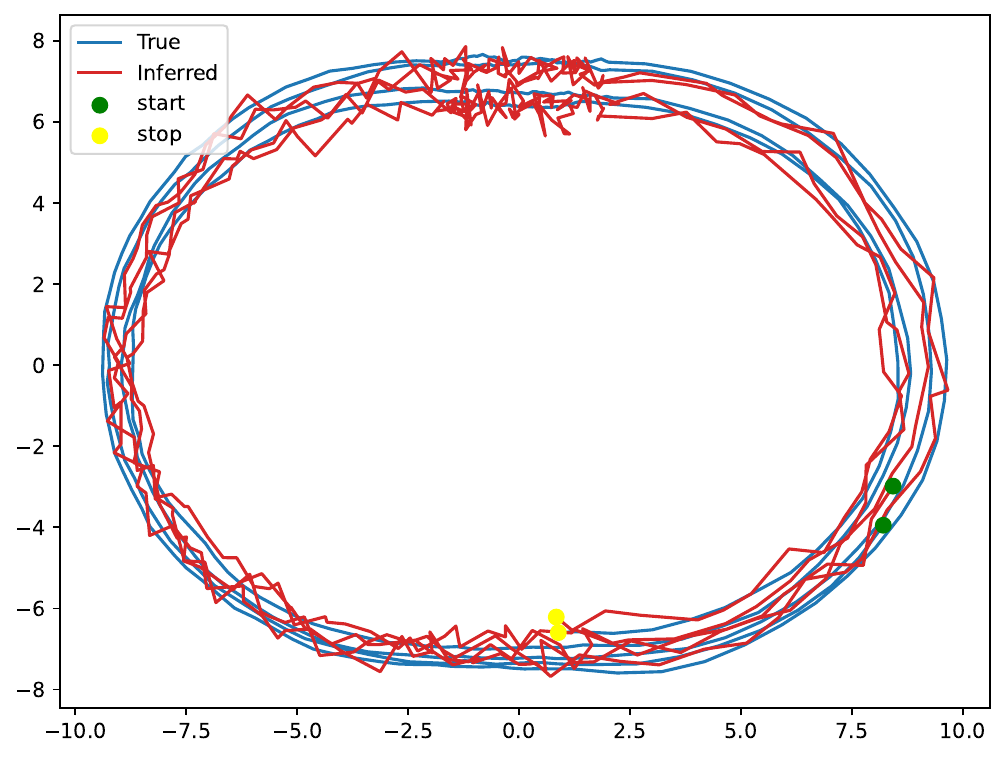}
        \label{fig:nascar_sub3}
    }

    \medskip

    \subfloat[RGPSSM (ours)]{%
        \includegraphics[width=0.3\textwidth]{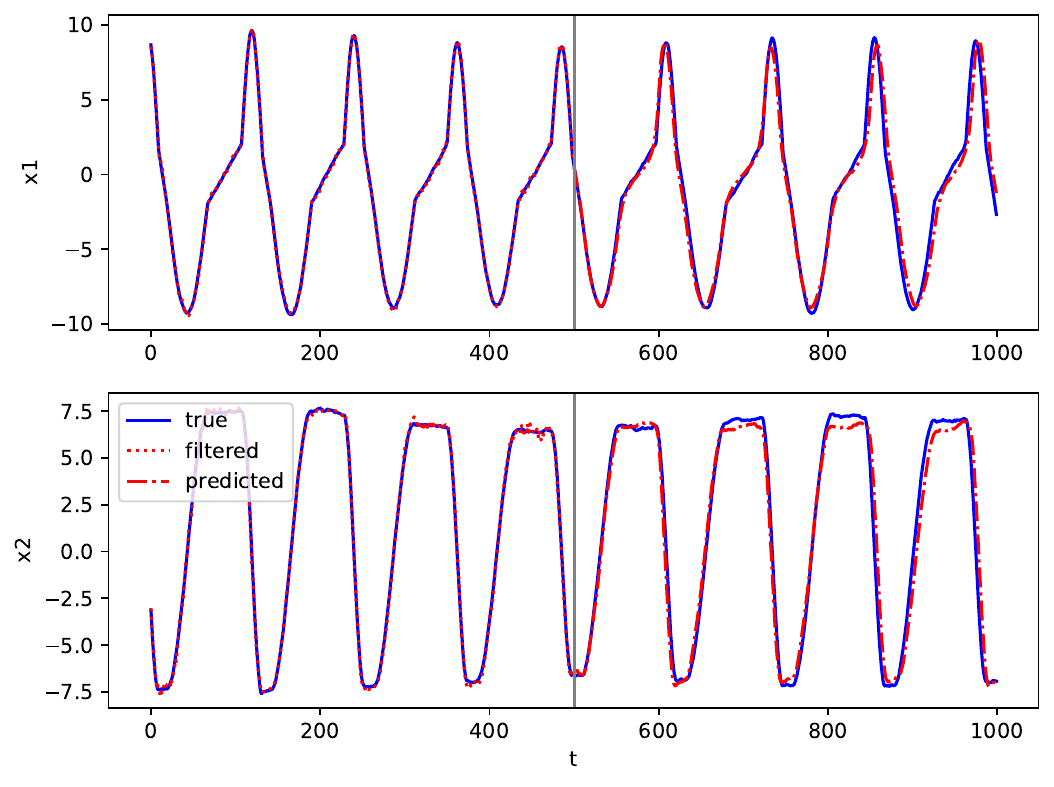}
        \label{fig:nascar_sub4}
    }
    \hspace{0.005\textwidth} 
    \subfloat[SVMC \cite{zhaoStreamingVariationalMonte2023}]{%
        \includegraphics[width=0.3\textwidth]{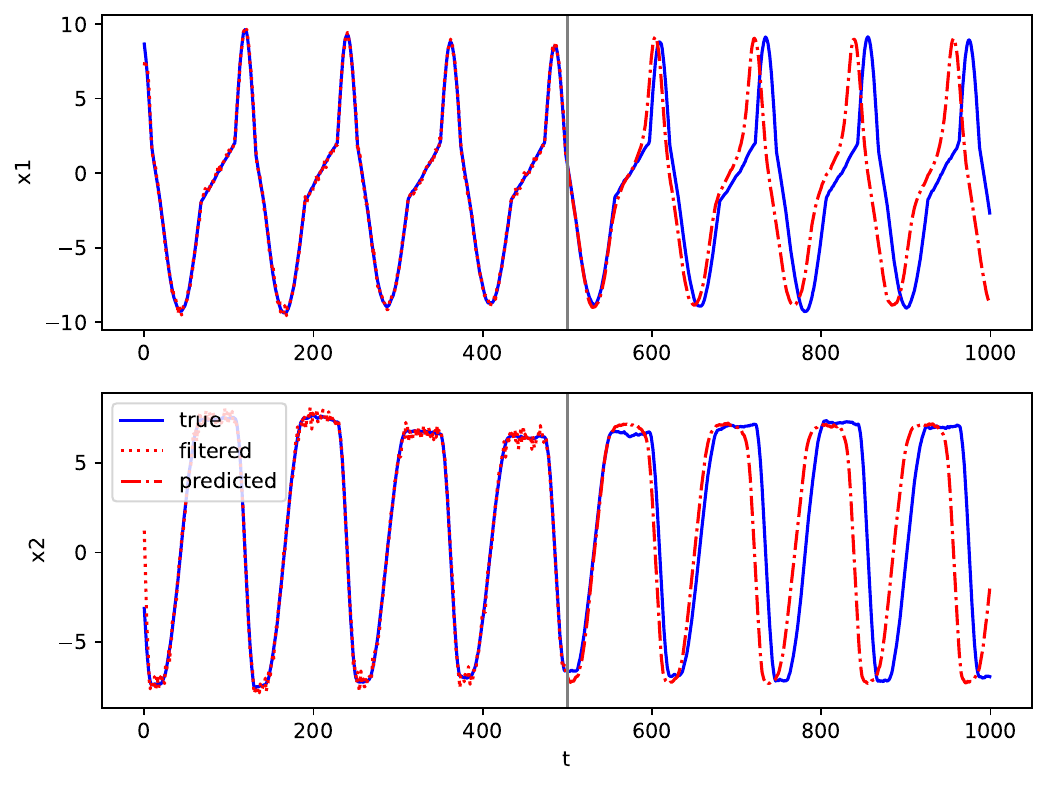}
        \label{fig:nascar_sub5}
    }
    \hspace{0.005\textwidth}
    \subfloat[OEnVI \cite{Lin2024}]{%
        \includegraphics[width=0.3\textwidth]{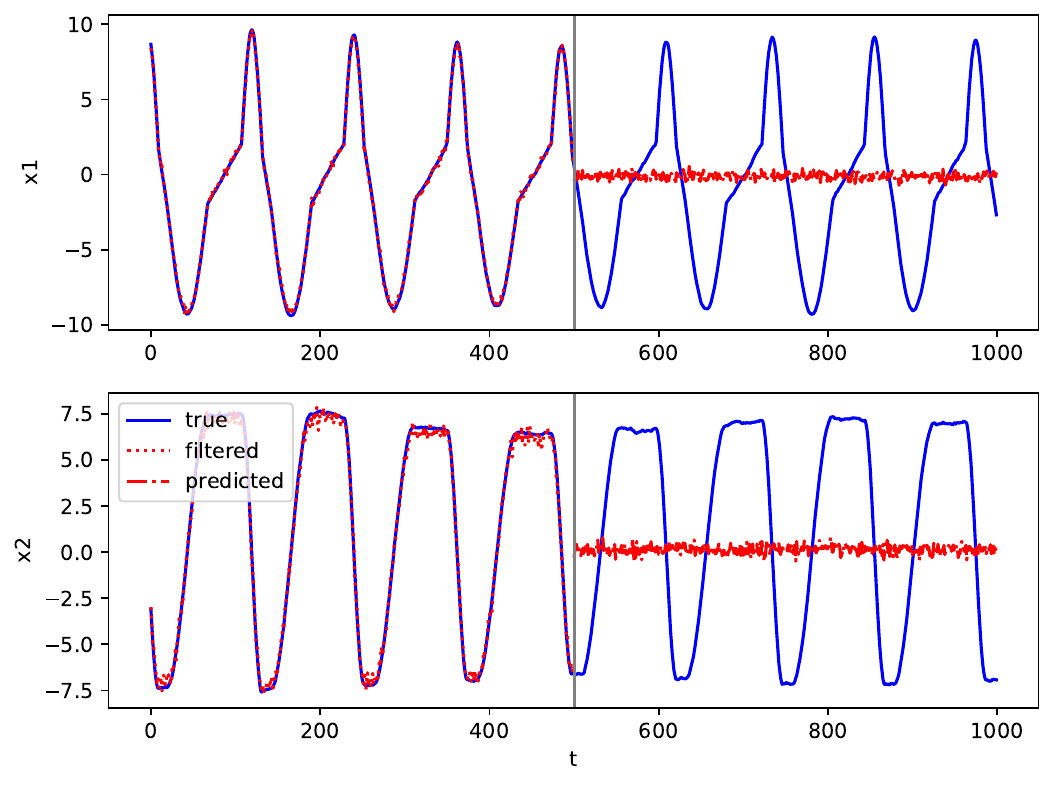}
        \label{fig:nascar_sub6}
    }

    \caption{Online learning results for NASCAR$^\circledR$ dynamics. Top: filtered state trajectory; Bottom: filtering and prediction.}
    \label{fig:nascar}
\end{figure*}

The experimental results are presented in Fig. \ref{fig:nascar}, where the top row showcases the true and filtered state trajectories, and the bottom row depicts the filtering and prediction results. It is observed that RGPSSM and SVMC effectively extract the dynamics and provide high-accuracy predictions, while OEnVI fails to learn the dynamics due to slow convergence and the large data requirements associated with SVI. Additionally, the learning accuracy and computational efficiency are quantified in Table \ref{tab:nascar}. As shown in the table, although SVMC is capable of learning system dynamics from limited measurements, it incurs a high computational cost due to the large number of particles required. In contrast, OEnVI achieves the lowest computation time but exhibits a slower learning convergence rate, primarily because it relies on a stochastic gradient optimization. 
However, the proposed method, grounded in the AKF framework, achieves an effective trade-off between learning convergence and computational efficiency. 
The EKF-like update mechanism provides second-order optimization speed~\cite{eon1998online}, which is generally faster than SVI-based approaches. These results highlight the advantages and superior performance of the proposed RGPSSM method.

\begin{table}[h]
\centering
\caption{Prediction Accuracy and Computational Efficiency of the Three Online GPSSM Methods.}
\begin{tabular}{cccccc}
\toprule
\textbf{Method} & \textbf{RGPSSM} & \textbf{SVMC} & \textbf{OEnVI} \\
\midrule
\textbf{Prediction RMSE} & \textbf{1.2552} & 5.9180 & 7.8583 \\
\textbf{Running Time (s)} & 8.59 & 42.69 & \textbf{6.55} \\
\bottomrule
\end{tabular}
\label{tab:nascar}
\end{table}

\subsection{Synthetic Wing Rock Dynamics Learning}\label{subsec:exp2}

To further evaluate RGPSSM's ability to online adapt GP hyperparameters and inducing points, it is tested on the synthetic wing rock dynamics learning task. In aerospace engineering, wing rock dynamics can cause roll oscillations in slender delta-winged aircraft, posing a significant threat to flight safety. To address this issue, most approaches rely on online learning of wing rock dynamics \cite{lee2014noncertainty,li2022nonlinear,singh1995direct,GirishChowdhary2015}. The associated dynamical system can be expressed by the following continuous-time SSM:

\begin{equation}\label{eq:wr_sys}
\begin{aligned}
\dot \theta &= p \\
\dot p &= L_{\delta_a} \delta_a + \Delta(\theta, p) 
\end{aligned}
\end{equation}
where \( \theta \in \mathbb{R} \) (\si{deg}) and \( p \in \mathbb{R} \) (\si{deg/\second}) denote the roll angle and roll rate, respectively.
As illustrated in \eqref{eq:wr_sys}, the dynamics model consists of a known part \( L_{\delta_a} \delta_a \), where $\delta_a$ is the aileron control input and \( L_{\delta_a} = 3~\si{\per\second\squared} \), and an unknown uncertainty model \( \Delta(\theta, p) \). The uncertainty model used in the simulation is taken from \cite{GirishChowdhary2015}, namely:

\begin{equation}
\begin{aligned}
\Delta(\theta,p) &= W_0 + W_1 \theta + W_2 p 
+ W_3 \vert \theta \vert p + W_4 \vert p \vert p + W_5 \theta^3
\end{aligned}
\end{equation}
where 
$W_0 = 0.8\,\si{deg \cdot s^{-2}}$, 
$W_1 = 0.2314\,\si{s^{-2}}$, 
$W_2 = 0.6918\,\si{s^{-1}}$, 
$W_3 = -0.6245\,\si{deg^{-1}s^{-1}}$, 
$W_4 = 0.0095\,\si{deg^{-1}}$, 
$W_5 = 0.0214\,\si{deg^{-2}\,s^{-2}}$.
To further demonstrate the learning capacity of the proposed method under sparse sensor conditions, the measurement is limited to the roll angle \( \theta \), which is corrupted by Gaussian white noise with a standard deviation of 0.2 degrees.

To apply the proposed Algorithm \ref{alg:rgpssm} to learn the uncertainty model $\Delta(\theta, p)$, the continuous-time dynamics \eqref{eq:wr_sys} is discretized with time step $\Delta t = 0.05\,\mathrm{s}$, resulting in the following transition model:

\begin{equation}
\begin{aligned}
\bm F \left(\bm x_t, c_t, f(\bm x_t \bm) \right) &= 
\bm x_t + 
\begin{bmatrix}
    \bm x_t^{(2)} \\
    f(\bm x_t) + L_{\delta_a} c_t  \\
\end{bmatrix} \Delta t
\end{aligned}
\end{equation}
where the system state is defined as $\bm x_t = [\theta(t),\, p(t)]^\top$, with $\bm x_t^{(2)} = p(t)$, the control input $c_t = \delta_a(t)$, and the uncertainty is modeled as $f(\bm x_t) = \Delta(\theta(t), p(t))$. To enable hyperparameter adaptation, we perform one iteration of hyperparameter optimization during each RGPSSM update using a learning rate of 0.01. 
The number of inducing points is limited to 20. Under these settings, we evaluate the proposed method and analyze the impact of adapting to the hyperparameter and inducing point. 

\begin{figure}[H]
\centering
\includegraphics[scale=0.5]{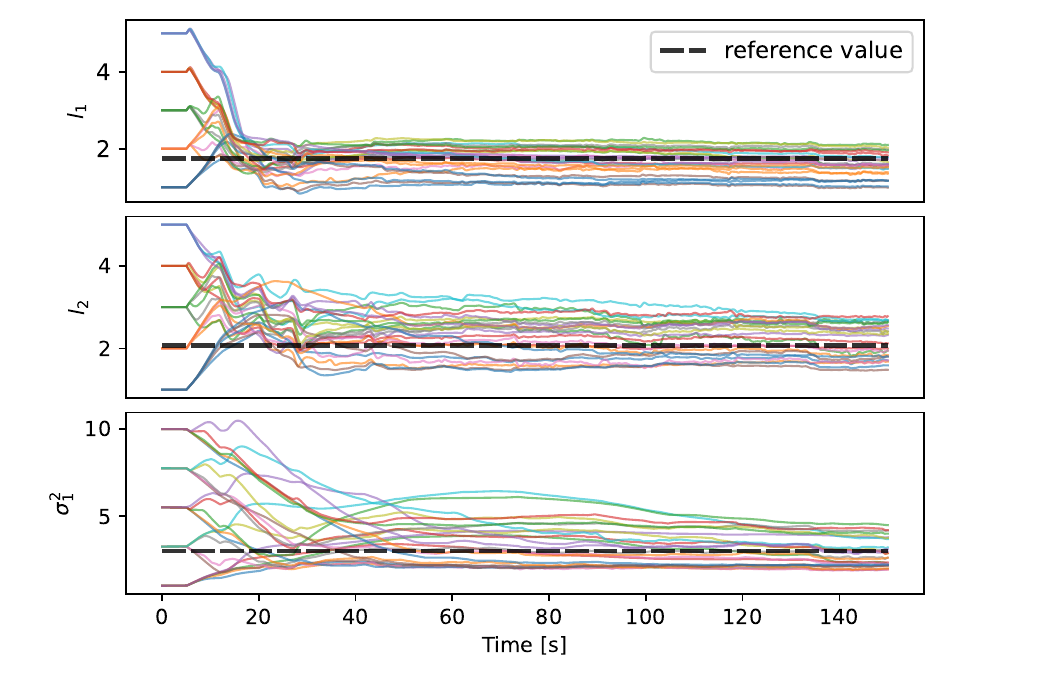}
\caption{Evolution of the GP hyperparameters during RGPSSM learning}
\label{fig:wr_hp}
\end{figure}
\begin{figure}[H]
\centering
\includegraphics[scale=0.5]{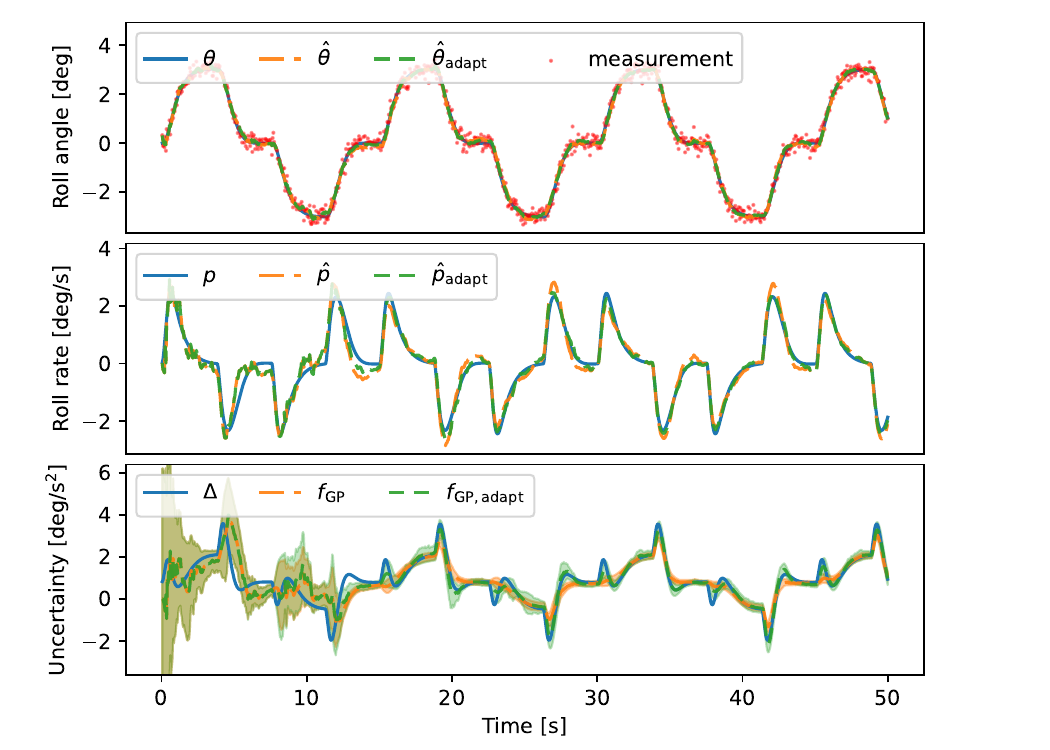}
\caption{Evolution of the system states and their estimation, where the initial values of the GP hyperparameters are \( l_1 = l_2 = 5 \) and \( \sigma_1^2 = 10 \). The shadows indicate 95\% confidence intervals.}
\label{fig:wr_state}
\end{figure}

Firstly, we conduct a Monte Carlo test of the proposed learning algorithm with different initial values of GP hyperparameters. The simulation results are depicted in Fig. \ref{fig:wr_hp} and \ref{fig:wr_state}. 
As shown in Fig. \ref{fig:wr_hp}, all three GP hyperparameters gradually converge near the reference values, which are obtained through offline GP training using data pairs $(\theta, p, \Delta)$. This indicates that the GP hyperparameters do not need to be precisely selected at the stage of algorithm initialization, and the online optimization can effectively adjust them. 
To evaluate the improvement resulting from hyperparameter adaptation, we calculate the prediction RMSE over the last quarter segment, which is 29.3\% lower with hyperparameter adaptation than without.
Furthermore, to more intuitively illustrate the benefits of hyperparameter adaptation, we present the state profile from one Monte Carlo simulation case, as shown in Fig. \ref{fig:wr_state}. It is evident that hyperparameter adaptation improves both the mean and variance predictions of the GP, thereby enhancing the filtering accuracy of the state, especially the roll rate \( p \). Therefore, the proposed method can not only learn the uncertain dynamics but also improve the state estimation accuracy.

\begin{figure*}[h]
\centering
\includegraphics[scale=0.44]{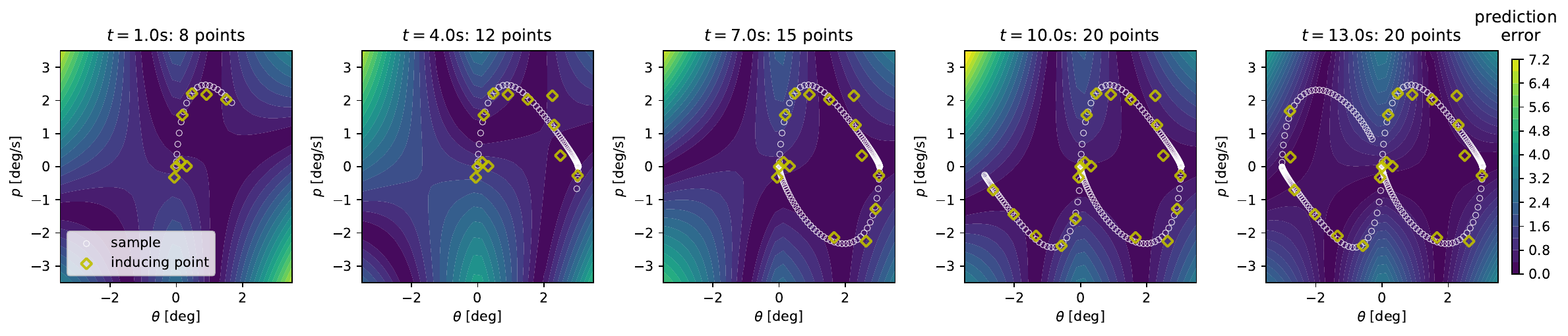}
\caption{Evolution of inducing-point set and prediction error over time.}
\label{fig:wr_id}
\end{figure*}

Secondly, to illustrate the adaptation process of the inducing points and their advantages, we present the associated results in Fig. \ref{fig:wr_id}. In this figure, the inducing inputs are represented by yellow diamonds, while the true state values are shown as white circles. It is evident that the number of inducing points gradually increases to the predefined budget ($M = 20$), and the points are concentrated around the existing states. Note that the inducing points are selected from state estimation values, and therefore do not coincide with the true states.
Given the localized nature of the squared exponential kernel, the selection of inducing points can effectively represent the GP model. This effectiveness is demonstrated in the background contour plot, which visualizes the distribution of prediction errors in the state space. It is evident that the prediction error decreases over time, particularly around the existing states, benefiting from the effective selection of inducing points.

\subsection{Real-World System Identification Task}\label{subsec:exp3}

To evaluate the generalization capability of the proposed method, we conducted experiments 
on five real-world system identification benchmark datasets\footnote{\url{https://homes.esat.kuleuven.be/~smc/daisy/daisydata.html}}.
 The first half of each dataset was used for training, while the second half was reserved for testing predictive performance. Due to limited prior knowledge of the models and the restricted amount of training data, these datasets are mainly used for assessing offline GPSSM approaches. In this study, we performed an online learning experiment with the proposed method, comparing its performance against three state-of-the-art offline GPSSM techniques: PRSSM \cite{doerrProbabilisticRecurrentStateSpace2018}, VCDT \cite{Ialongo2019}, and EnVI \cite{Lin2024}. 
For each method, the number of inducing points was capped at 20, and a 4-dimensional state GPSSM was used to capture system dynamics, with a transition model defined as \( \bm{x}_{t+1} = f(\bm{x}_t, c_{t+1}) \), where \( \bm{x}_t \in \mathbb{R}^4 \) and $c_{t+1}$ is the control input, and a measurement model given by \( y_t = [1, 0, 0, 0] \bm{x}_t \). 
In addition, to activate the estimation for the latent state within RGPSSM, we randomly assign an inducing point around the origin before learning to make the transition Jacobian \( A_x \) non-zero for each state dimension (see our code for details).

The experimental results from multiple simulations with different random seeds are summarized in Table \ref{tab:sysID}. As shown in the table, although online learning is challenging, the proposed method achieves the best prediction accuracy on three datasets. This superior performance can be attributed to the fact that while offline methods can utilize the full dataset for repeated training, their learning accuracy is often limited by various factors, such as local optima, vanishing or exploding gradients during SSM/RNN training \cite{pascanu2013difficulty}, or insufficient expressiveness of the variational distribution (e.g., the mean-field assumption in PRSSM \cite{doerrProbabilisticRecurrentStateSpace2018}, Gaussian parametric constraints over the state and inducing points posterior in VCDT \cite{Ialongo2019}, or using the filtering distribution to approximate the smoothing distribution in EnVI \cite{Lin2024}). However, RGPSSM fully considers the coupling between the system state and the GP, with only the linearization approximation limiting its learning capability.
Therefore, in some cases, the online learning performance of the proposed method is comparable to that of offline methods. Of course, due to the linearization approximation, the proposed method may have limitations in highly nonlinear scenarios, which could potentially be addressed by integrating advanced nonlinear filtering techniques. 
Additionally, we mention that the results of online learning have the potential to provide effective initializations for offline learning, thereby improving both training speed and accuracy. This is an interesting direction that can be further explored in future work.


\begin{table}[h]
    \centering
    \caption{Performance Comparison of Different GPSSM Methods across Five System Identification Benchmarks. Prediction RMSE is Averaged over Five Seeds, with Standard Deviations Reported in Parentheses.}
    \resizebox{\columnwidth}{!}{%
    \begin{tabular}{cccccc}
    \toprule
    \multirow{2}{*}[-2pt]{\textbf{Methods}} & \multicolumn{3}{c}{\textbf{Offline}} & \textbf{Online} \\ 
    \cmidrule(lr){2-4} \cmidrule(lr){5-5}
     & \textbf{PRSSM} & \textbf{VCDT} & \textbf{EnVI} & \textbf{RGPSSM} \\ 
    \midrule
    \textbf{Actuator} & \makecell{0.691 \\ (0.148)} & \makecell{0.815 \\ (0.012)} & \makecell{\textbf{0.657} \\ \textbf{(0.095)}} & \makecell{1.048 \\ (0.142)} \\
    \midrule
    \textbf{Ball Beam} & \makecell{0.074 \\ (0.010)} & \makecell{0.065 \\ (0.005)} & \makecell{0.055 \\ (0.002)} & \makecell{\textbf{0.046} \\ \textbf{(0.002)}} \\
    \midrule
    \textbf{Drive} & \makecell{\textbf{0.647} \\ \textbf{(0.057)}} & \makecell{0.735 \\ (0.005)} & \makecell{0.703 \\ (0.050)} & \makecell{0.863 \\ (0.083)} \\
    \midrule
    \textbf{Dryer} & \makecell{0.174 \\ (0.013)} & \makecell{0.667 \\ (0.266)} & \makecell{0.125 \\ (0.017)} & \makecell{\textbf{0.105} \\ \textbf{(0.004)}} \\
    \midrule
    \textbf{Gas Furnace} & \makecell{1.503 \\ (0.196)} & \makecell{2.052 \\ (0.163)} & \makecell{1.388 \\ (0.123)} & \makecell{\textbf{1.373} \\ \textbf{(0.061)}} \\
    \bottomrule
    \end{tabular}%
    }
    \label{tab:sysID}
\end{table}

\section{Conclusion}

In this paper, we propose a novel recursive learning method, RGPSSM, for online inference in Gaussian Process State Space Models (GPSSMs). First, to enable learning in any operating domain, we derive a two-step Bayesian update equation that does not rely on any GP approximation, which provides a new solution for the nonlinearity and nonparametric issues of GPSSMs.
Second, to mitigate the computational burden imposed by the nonparametric nature of GPs, we introduce a dynamic inducing-point set adjustment algorithm to ensure computational efficiency.  
Third, to eliminate the need for careful tuning of GP hyperparameters at initialization and to enhance online learning accuracy, we develop an online optimization method that extracts measurement information from the approximate posterior distribution.
Fourth, we derive a Cholesky version of RGPSSM in detail to further improve numerical stability, and also present the extension of RGPSSM to the multi-output GP case.
Fifth, we analyze the computational complexity of the proposed method and demonstrate that it generalizes several existing approaches, as well as highlight its strong potential for further extensions.
Finally, through extensive evaluations on a diverse set of real and synthetic datasets, we demonstrate the superior learning performance and adaptability of the proposed method. 

\section{Acknowledgement}
This work is supported by the National Natural Science Foundation of China (Grant No. 12572052) and the Beijing Natural Science Foundation (Grant L251013).

\bibliographystyle{unsrt}        
\bibliography{reference}           

\begin{thebibliography}{10}

\bibitem{he2025novel}
Xiaodong He, Weijia Yao, Zhiyong Sun, and Zhongkui Li.
\newblock A novel vector-field-based motion planning algorithm for 3d
  nonholonomic robots.
\newblock {\em Automatica}, 172:111996, 2025.

\bibitem{kang2026stochastic}
Chang~Ho Kang and Sun~Young Kim.
\newblock Stochastic gradient hamiltonian sequential monte carlo filter with
  earth mover’s distance sampling for target tracking.
\newblock {\em Automatica}, 183:112599, 2026.

\bibitem{zhang2023group}
Xuqi Zhang, Haiqi Liu, Fanqin Meng, and Xiaojing Shen.
\newblock Group target tracking via jointly optimizing group partition and
  association.
\newblock {\em Automatica}, 153:111013, 2023.

\bibitem{breschi2023data}
Valentina Breschi, Alessandro Chiuso, and Simone Formentin.
\newblock Data-driven predictive control in a stochastic setting: a unified
  framework.
\newblock {\em Automatica}, 152:110961, 2023.

\bibitem{kessler2026design}
Nicolas Kessler and Lorenzo Fagiano.
\newblock On the design of linear time varying model predictive control for
  trajectory stabilization.
\newblock {\em Automatica}, 183:112577, 2026.

\bibitem{da2025fault}
M{\'a}rcia~Luciana da~Costa~Peixoto, Pedro~Moreira de~Oliveira, Iury Bessa,
  Pedro~Henrique Coutinho, Paulo Sergio~Pereira Pessim, Vicenc Puig, and
  Reinaldo~Martinez Palhares.
\newblock Fault-tolerant dynamic output feedback control of lpv systems via
  fault hiding.
\newblock {\em Automatica}, 174:112191, 2025.

\bibitem{2025Experience}
Chaoran Qu, Lin Cheng, Shengping Gong, and Xu~Huang.
\newblock Experience replay enhances excitation condition of neural-network
  adaptive control learning.
\newblock {\em Journal of Guidance, Control and Dynamics}, 48(3):496--507,
  2025.

\bibitem{Berntorp2021}
Karl Berntorp.
\newblock Online bayesian inference and learning of gaussian-process
  state--space models.
\newblock {\em Automatica}, 129:109613, 2021.

\bibitem{rangapuram2018deep}
Syama~Sundar Rangapuram, Matthias~W Seeger, Jan Gasthaus, Lorenzo Stella,
  Yuyang Wang, and Tim Januschowski.
\newblock Deep state space models for time series forecasting.
\newblock {\em Advances in neural information processing systems}, 31, 2018.

\bibitem{gedon2021deep}
Daniel Gedon, Niklas Wahlstr{\"o}m, Thomas~B Sch{\"o}n, and Lennart Ljung.
\newblock Deep state space models for nonlinear system identification.
\newblock {\em IFAC-PapersOnLine}, 54(7):481--486, 2021.

\bibitem{frigolaBayesianTimeSeries2015}
Roger Frigola-Alcalde.
\newblock {\em Bayesian time series learning with Gaussian processes}.
\newblock Ph.d. thesis, University of Cambridge, Cambridge, UK, August 2015.

\bibitem{zheng2025model}
Tengjie Zheng, Lin Cheng, Shengping Gong, and Xu~Huang.
\newblock Model incremental learning of flight dynamics enhanced by sample
  management.
\newblock {\em Aerospace Science and Technology}, page 110049, 2025.

\bibitem{turner2010state}
Ryan Turner, Marc Deisenroth, and Carl Rasmussen.
\newblock State-space inference and learning with gaussian processes.
\newblock In {\em Proceedings of the thirteenth international conference on
  artificial intelligence and statistics}, pages 868--875. JMLR Workshop and
  Conference Proceedings, 2010.

\bibitem{rasmussenGaussianProcessesMachine2008}
Carl~Edward Rasmussen and Christopher K.~I. Williams.
\newblock Gaussian processes for machine learning.
\newblock In {\em Gaussian Processes for Machine Learning}, chapter~2. MIT
  Press, Cambridge, MA, USA, 2006.

\bibitem{pillonetto2022regularization}
Gianluigi Pillonetto, Tianshi Chen, Alessandro Chiuso, Giuseppe De~Nicolao, and
  Lennart Ljung.
\newblock Regularization for nonlinear system identification.
\newblock In {\em Regularized System Identification: Learning Dynamic Models
  from Data}, pages 313--342. Springer, 2022.

\bibitem{care2023kernel}
Algo Car{\`e}, Ruggero Carli, Alberto Dalla~Libera, Diego Romeres, and
  Gianluigi Pillonetto.
\newblock Kernel methods and gaussian processes for system identification and
  control: A road map on regularized kernel-based learning for control.
\newblock {\em IEEE Control Systems Magazine}, 43(5):69--110, 2023.

\bibitem{fisac2018general}
Jaime~F Fisac, Anayo~K Akametalu, Melanie~N Zeilinger, Shahab Kaynama, Jeremy
  Gillula, and Claire~J Tomlin.
\newblock A general safety framework for learning-based control in uncertain
  robotic systems.
\newblock {\em IEEE Transactions on Automatic Control}, 64(7):2737--2752, 2018.

\bibitem{frigola2013bayesian}
Roger Frigola, Fredrik Lindsten, Thomas~B Sch{\"o}n, and Carl~Edward Rasmussen.
\newblock Bayesian inference and learning in gaussian process state-space
  models with particle mcmc.
\newblock {\em Advances in neural information processing systems}, 26, 2013.

\bibitem{frigola2014variational}
Roger Frigola, Yutian Chen, and Carl~Edward Rasmussen.
\newblock Variational gaussian process state-space models.
\newblock {\em Advances in neural information processing systems}, 27, 2014.

\bibitem{svensson2017flexible}
Andreas Svensson and Thomas~B Sch{\"o}n.
\newblock A flexible state--space model for learning nonlinear dynamical
  systems.
\newblock {\em Automatica}, 80:189--199, 2017.

\bibitem{fan2023free}
Xuhui Fan, Edwin~V Bonilla, Terence O’Kane, and Scott~A Sisson.
\newblock Free-form variational inference for gaussian process state-space
  models.
\newblock In {\em International Conference on Machine Learning}, pages
  9603--9622. PMLR, 2023.

\bibitem{eleftheriadis2017identification}
Stefanos Eleftheriadis, Tom Nicholson, Marc Deisenroth, and James Hensman.
\newblock Identification of gaussian process state space models.
\newblock {\em Advances in neural information processing systems}, 30, 2017.

\bibitem{doerrProbabilisticRecurrentStateSpace2018}
Andreas Doerr, Christian Daniel, Martin Schiegg, Nguyen-Tuong Duy, Stefan
  Schaal, Marc Toussaint, and Trimpe Sebastian.
\newblock Probabilistic recurrent state-space models.
\newblock In {\em International conference on machine learning}, pages
  1280--1289. PMLR, 2018.

\bibitem{Ialongo2019}
Alessandro~Davide Ialongo, Mark Van Der~Wilk, James Hensman, and Carl~Edward
  Rasmussen.
\newblock Overcoming mean-field approximations in recurrent gaussian process
  models.
\newblock In {\em International Conference on Machine Learning}, pages
  2931--2940. PMLR, 2019.

\bibitem{lindinger2022laplace}
Jakob Lindinger, Barbara Rakitsch, and Christoph Lippert.
\newblock Laplace approximated gaussian process state-space models.
\newblock In {\em Uncertainty in Artificial Intelligence}, pages 1199--1209.
  PMLR, 2022.

\bibitem{Lin2024}
Zhidi Lin, Yiyong Sun, Feng Yin, and Alexandre~Hoang Thi{\'e}ry.
\newblock Ensemble kalman filtering meets gaussian process ssm for
  non-mean-field and online inference.
\newblock {\em IEEE Transactions on Signal Processing}, 2024.

\bibitem{veiback2019learning}
Clas Veib{\"a}ck, Jonatan Olofsson, Tom~Rune Lauknes, and Gustaf Hendeby.
\newblock Learning target dynamics while tracking using gaussian processes.
\newblock {\em IEEE Transactions on Aerospace and Electronic Systems},
  56(4):2591--2602, 2019.

\bibitem{Kullberg2020}
Anton Kullberg, Isaac Skog, and Gustaf Hendeby.
\newblock Learning driver behaviors using a gaussian process augmented
  state-space model.
\newblock In {\em 2020 IEEE 23rd International Conference on Information Fusion
  (FUSION)}, pages 1--7. IEEE, 2020.

\bibitem{Kullberg2021}
Anton Kullberg.
\newblock {\em On Joint State Estimation and Model Learning Using Gaussian
  Process Approximations}.
\newblock PhD thesis, Link{\"o}ping University Electronic Press, 2021.

\bibitem{zhaoStreamingVariationalMonte2023}
Yuan Zhao, Josue Nassar, Ian Jordan, M{\'o}nica Bugallo, and Il~Memming Park.
\newblock Streaming variational monte carlo.
\newblock {\em IEEE transactions on pattern analysis and machine intelligence},
  45(1):1150--1161, 2022.

\bibitem{liu2023sequential}
Yuhao Liu, Marzieh Ajirak, and Petar~M Djuri{\'c}.
\newblock Sequential estimation of gaussian process-based deep state-space
  models.
\newblock {\em IEEE Transactions on Signal Processing}, 2023.

\bibitem{grisetti2007improved}
Giorgio Grisetti, Cyrill Stachniss, and Wolfram Burgard.
\newblock Improved techniques for grid mapping with rao-blackwellized particle
  filters.
\newblock {\em IEEE transactions on Robotics}, 23(1):34--46, 2007.

\bibitem{doucet2009tutorial}
Arnaud Doucet, Adam~M Johansen, et~al.
\newblock A tutorial on particle filtering and smoothing: Fifteen years later.
\newblock {\em Handbook of nonlinear filtering}, 12(656-704):3, 2009.

\bibitem{Park2022}
Soon-Seo Park, Young-Jin Park, Youngjae Min, and Han-Lim Choi.
\newblock Online gaussian process state-space model: Learning and planning for
  partially observable dynamical systems.
\newblock {\em International Journal of Control, Automation and Systems},
  20(2):601--617, 2022.

\bibitem{ljung2003asymptotic}
Lennart Ljung.
\newblock Asymptotic behavior of the extended kalman filter as a parameter
  estimator for linear systems.
\newblock {\em IEEE Transactions on Automatic Control}, 24(1):36--50, 2003.

\bibitem{oconnellNeuralFlyEnablesRapid2022}
Michael O’Connell, Guanya Shi, Xichen Shi, Kamyar Azizzadenesheli, Anima
  Anandkumar, Yisong Yue, and Soon-Jo Chung.
\newblock Neural-fly enables rapid learning for agile flight in strong winds.
\newblock {\em Science Robotics}, 7(66):eabm6597, 2022.

\bibitem{VanVaerenbergh2012}
Steven Van~Vaerenbergh, Miguel L{\'a}zaro-Gredilla, and Ignacio
  Santamar{\'\i}a.
\newblock Kernel recursive least-squares tracker for time-varying regression.
\newblock {\em IEEE transactions on neural networks and learning systems},
  23(8):1313--1326, 2012.

\bibitem{csatoSparseLineGaussian2002}
Lehel Csat{\'o} and Manfred Opper.
\newblock Sparse on-line gaussian processes.
\newblock {\em Neural computation}, 14(3):641--668, 2002.

\bibitem{csatoGaussianProcessesIterative2002}
Lehel Csat{\'o}.
\newblock {\em Gaussian processes: iterative sparse approximations}.
\newblock PhD thesis, Aston University, Birmingham, UK, March 2002.

\bibitem{buiStreamingSparseGaussian2017}
Thang~D Bui, Cuong Nguyen, and Richard~E Turner.
\newblock Streaming sparse gaussian process approximations.
\newblock {\em Advances in Neural Information Processing Systems}, 30, 2017.

\bibitem{kingma2014adam}
Diederik~P Kingma.
\newblock Adam: A method for stochastic optimization.
\newblock {\em arXiv preprint arXiv:1412.6980}, 2014.

\bibitem{alvarez2012kernels}
Mauricio~A Alvarez, Lorenzo Rosasco, Neil~D Lawrence, et~al.
\newblock Kernels for vector-valued functions: A review.
\newblock {\em Foundations and Trends{\textregistered} in Machine Learning},
  4(3):195--266, 2012.

\bibitem{eon1998online}
L~Eon~Bottou.
\newblock Online learning and stochastic approximations.
\newblock {\em Online learning in neural networks}, 17(9):142, 1998.

\bibitem{singh1995direct}
Sahjendra~N Singh, Woosoon Yirn, and William~R Wells.
\newblock Direct adaptive and neural control of wing-rock motion of slender
  delta wings.
\newblock {\em Journal of Guidance, control, and Dynamics}, 18(1):25--30, 1995.

\bibitem{wan2000unscented}
Eric~A Wan and Rudolph Van Der~Merwe.
\newblock The unscented kalman filter for nonlinear estimation.
\newblock In {\em Proceedings of the IEEE 2000 adaptive systems for signal
  processing, communications, and control symposium (Cat. No. 00EX373)}, pages
  153--158. Ieee, 2000.

\bibitem{arasaratnam2009cubature}
Ienkaran Arasaratnam and Simon Haykin.
\newblock Cubature kalman filters.
\newblock {\em IEEE Transactions on automatic control}, 54(6):1254--1269, 2009.

\bibitem{kullberg2024dynamically}
Anton Kullberg, Martin~A Skoglund, Isaac Skog, and Gustaf Hendeby.
\newblock Dynamically iterated filters: A unified framework for improved
  iterated filtering and smoothing.
\newblock {\em arXiv preprint arXiv:2404.15359}, 2024.

\bibitem{dowling2024exponential}
Matthew Dowling, Yuan Zhao, and Memming Park.
\newblock exponential family dynamical systems (xfads): Large-scale nonlinear
  gaussian state-space modeling.
\newblock {\em Advances in Neural Information Processing Systems},
  37:13458--13488, 2024.

\bibitem{deisenroth2009analytic}
Marc~Peter Deisenroth, Marco~F Huber, and Uwe~D Hanebeck.
\newblock Analytic moment-based gaussian process filtering.
\newblock In {\em Proceedings of the 26th annual international conference on
  machine learning}, pages 225--232, 2009.

\bibitem{lindermanBayesianLearningInference2017}
Scott Linderman, Matthew Johnson, Andrew Miller, Ryan Adams, David Blei, and
  Liam Paninski.
\newblock Bayesian learning and inference in recurrent switching linear
  dynamical systems.
\newblock In {\em Artificial intelligence and statistics}, pages 914--922.
  PMLR, 2017.

\bibitem{lee2014noncertainty}
Keum~W Lee and Sahjendra~N Singh.
\newblock Noncertainty-equivalent adaptive wing-rock control via chebyshev
  neural network.
\newblock {\em Journal of Guidance, Control, and Dynamics}, 37(1):123--133,
  2014.

\bibitem{li2022nonlinear}
Dongyang Li, Antonios Tsourdos, Zhongyuan Wang, and Dmitry Ignatyev.
\newblock Nonlinear analysis for wing-rock system with adaptive control.
\newblock {\em Journal of Guidance, Control, and Dynamics}, 45(11):2174--2181,
  2022.

\bibitem{GirishChowdhary2015}
Girish Chowdhary, Hassan~A Kingravi, Jonathan~P How, and Patricio~A Vela.
\newblock Bayesian nonparametric adaptive control using gaussian processes.
\newblock {\em IEEE transactions on neural networks and learning systems},
  26(3):537--550, 2014.

\bibitem{pascanu2013difficulty}
Razvan Pascanu, Tomas Mikolov, and Yoshua Bengio.
\newblock On the difficulty of training recurrent neural networks.
\newblock In {\em International conference on machine learning}, pages
  1310--1318. Pmlr, 2013.

\bibitem{golubMatrixComputations2013}
Gene~H. Golub and Charles F.~Van Loan.
\newblock {\em Matrix Computations}, chapter~6, pages 352--355.
\newblock Johns Hopkins Studies in the Mathematical Sciences. JHU Press,
  Baltimore, 4 edition, 2013.
\newblock Section 6.5.4: Cholesky Updating and Downdating.

\end{thebibliography}

\appendix
\section{Score for Selection of Inducing Point}\label{app:score}

This section derives the score quantifying the accuracy loss for discarding an inducing point, which is achieved based on the KL divergence in \eqref{eq:Dstar}. As shown in \eqref{eq:Dstar}, $D_1$ is unrelated to the point to discard $u_{\mathrm{d}}$. Therefore, only $D_2$ and $D_3$ need to be evaluated. Without loss of generality, suppose the index of the discarded point $u_{\mathrm{d}}$ is $i_{\mathrm{d}} = 1$.

Firstly, for $D_2$, the log-term in the integral is:

\begin{equation}\label{eq:d2}
\begin{aligned}
\log p(u_{\mathrm{d}} | \bm{u}_{\mathrm{l}}) &= -\dfrac{1}{2} \log (2 \pi )
- \dfrac{1}{2} \log \vert \gamma_d \vert \\
&\,- \dfrac{1}{2} (\bm{u}_d - \bm{k}_{dl} \bm{u}_{\mathrm{l}}) ^T \gamma_d^{-1} (\bm{u}_d - \bm{k}_{dl} \bm{u}_{\mathrm{l}})
\end{aligned}
\end{equation}
where \( \bm{k}_{dl} = \bm{K}_{dl} \bm{K}_{ll}^{-1} \) and \( \gamma_d = \bm{K}_{dd} - \bm{K}_{dl} \bm{K}_{ll}^{-1} \bm{K}_{ld} \). Denote \( \bm{Q} = \bm{K}_{uu}^{-1} \), and using the matrix inversion formula (see Appendix A of \cite{csatoGaussianProcessesIterative2002}), we have:

\begin{equation}\label{eq:gam_kdl}
\begin{aligned}
\gamma_d &=  {\bm{Q}_{dd}}^{-1} \\
\bm{k}_{dl} &= - {\bm{Q}_{dd}}^{-1} \bm{Q}_{dl}
\end{aligned}
\end{equation}
where \( \bm{Q}_{dd} \) denotes the \( i_{\mathrm{d}} \)-th diagonal element of matrix \( \bm{Q} \), and \( \bm{Q}_{dl} \) denotes the \( i_{\mathrm{d}} \)-th row of \( \bm{Q} \) with \( \bm{Q}_{dd} \) removed.

Using \eqref{eq:Dstar}, \eqref{eq:d2} and \eqref{eq:gam_kdl}, the $u_{\mathrm{d}}$-relevant term in $D_2$ is:

\begin{equation}\label{eq:D2}
\begin{aligned}
\tilde D_2 
&= \dfrac{1}{2} \log \vert \bm{Q}_{dd} \vert - \dfrac{1}{2} \Delta \\
\Delta &= \int q(\bm{u}) (u_{\mathrm{d}} - \bm{k}_{dl} \bm{u}_{\mathrm{l}})^T \bm{Q}_{dd} (u_{\mathrm{d}} - \bm{k}_{dl} \bm{u}_{\mathrm{l}}) \, \mathrm{d}\bm{u} \\
&= \mathbb{E}_{q(\bm{u})} \left[ \mathrm{tr} 
\left( \bm{Q}_{dd} (u_{\mathrm{d}} - \bm{k}_{dl} \bm{u}_{\mathrm{l}}) 
(u_{\mathrm{d}} - \bm{k}_{dl} \bm{u}_{\mathrm{l}})^T \right) \right] \\
&=  \mathrm{tr} 
\left( \bm{Q}_{dd} \mathbb{E}_{q(\bm{u})} 
\left[ (u_{\mathrm{d}} - \bm{k}_{dl} \bm{u}_{\mathrm{l}}) (u_{\mathrm{d}} - \bm{k}_{dl} \bm{u}_{\mathrm{l}})^T \right]  \right)
\end{aligned}
\end{equation}

Let $\bm{\phi} = [1, -\bm{k}_{dl}]$, and then we have $u_{\mathrm{d}} - \bm{k}_{dl} \bm{u}_{\mathrm{l}} = \bm{\phi} \bm{u}$. Using \eqref{eq:gam_kdl}, we further have:

\begin{equation}
\begin{aligned}
\bm{\phi} & = \begin{bmatrix} 1 & {\bm{Q}}_{dd}^{-1} \bm{Q}_{dl} \\ \end{bmatrix} \\
&= \begin{bmatrix} {\bm{Q}}_{dd}^{-1} \bm{Q}_{dd} & {\bm{Q}}_{dd}^{-1} \bm{Q}_{dl} \\ \end{bmatrix} \\
&= {\bm{Q}}_{dd}^{-1} \bm{Q}_{du}
\end{aligned}
\end{equation}
where $\bm{Q}_{du}$ denotes the $i_{\mathrm{d}}$th row of $\bm{Q}$. Therefore:

\begin{equation}
\begin{aligned}
\Delta &= \mathrm{tr} 
\left( \bm{Q}_{dd} \mathbb{E}_{q(\bm{u})} \left[ \bm{\phi} \bm{u} \bm{u}^T \bm{\phi}^T \right]  \right) \\
&= \mathrm{tr} 
\left( \bm{Q}_{dd} \left[ \bm{\phi} (\bm{S}_{uu} + \bm{m}_{u} \bm{m}_{u}^T ) \bm{\phi}^T \right]  \right) \\
&= \Delta_1 + \Delta_2
\end{aligned}
\end{equation}
where
\begin{equation}\label{eq:2Delta}
\begin{aligned}
\Delta_1 &= \bm{m}_u^T \bm{Q}_{du}^T {\bm{Q}}_{dd}^{-1} \bm{Q}_{du} \bm{m}_u \\
\Delta_2 &= \mathrm{tr}\left(\bm{Q}_{du} \bm{S}_{uu} \bm{Q}_{du}^T {\bm{Q}}_{dd}^{-1} \right)
\end{aligned}
\end{equation}

For $D_3$, we can use the differential entropy identical for Gaussian distribution:

\begin{equation}\label{eq:diff_entropy}
\begin{aligned}
\int p(\bm{a}) \log p(\bm{a}) \mathrm{d}\bm{a}= -\dfrac{1}{2} \left[ n_a + n_a\log(2\pi) + \log\vert \bm{\Sigma}_a \vert \right]
\end{aligned}
\end{equation}
where $p(\bm{a})$ is a $n_a$-dimension Gaussian distribution with covariance $\bm{\Sigma}_a$. 
Deonte the covariance of joint distribution $q(\bm{X}_t)$ that marginal out $u_{\mathrm{d}}$ as $\bm{\Sigma}_{t,l}$, and denote the inversion of original covariance as $\bm{\Omega} = \bm{\Sigma}_{t}^{-1}$. Then, given \eqref{eq:diff_entropy}, the $u_{\mathrm{d}}$ relevant term in $D_3$ is:

\begin{equation}\label{eq:D3}
\begin{aligned}
-\dfrac{1}{2} \log \vert \bm{\Sigma}_{t, l} \vert
&=  -\dfrac{1}{2} \log \left( \vert \bm{\Sigma}_t \vert / \vert \bm{S}_{dd} - \bm{\zeta}_d^T  \bm{\Sigma}_{t,l} \bm{\zeta}_d \vert \right) \\
&= -\dfrac{1}{2} \log  \vert \bm{\Sigma}_t \vert -\dfrac{1}{2} \log \vert \bm{\Omega}_{dd}  \vert  \\
\bm{\zeta}_d& = \begin{bmatrix}
\bm{V}_{xd} \\
\bm{S}_{ud} \\
\end{bmatrix}
\end{aligned}
\end{equation}
where $\bm{\Omega}_{dd}$ is the element of the matrix \( \bm{\Omega} \) corresponding to the discarded point \( u_{\mathrm{d}} \), namely, its \((d_x + i_{\mathrm{d}})\)-th diagonal element. In addition, \eqref{eq:D3} is derived by using the properties of the determinants of block matrices and the matrix inversion formula (see Appendix A of \cite{csatoGaussianProcessesIterative2002}).

Based on the KL divergence \eqref{eq:Dstar} and combing \eqref{eq:D2}, \eqref{eq:2Delta} and \eqref{eq:D3}, we have the score:

\begin{equation}
\begin{aligned}
s_d = \Delta_1 + \Delta_2 + \Delta_3 \\
\end{aligned}
\end{equation}
where

\begin{equation}
\Delta_3 = \log\vert \bm{\Omega}_{dd} \vert -\log\vert \bm{Q}_{dd} \vert
\end{equation}

In terms of effect, a lower value of the score $s_d$ implies less accuracy loss for discarding the point $u_{\mathrm{d}}$.

\section{Prediction Equation without Adding Points}\label{app:pred}

According to \eqref{eq:res_proj}, if the new inducing point $f_t$ is discarded in the prediction step, the distribution after discarding can be given by:

\begin{equation}
\begin{aligned}
\hat q^{*}(\bm{x}_{t+1}, \bm{u})
&= \int q^-(\bm{x}_{t+1}, \bar {\bm{u}}) \mathrm{d} f_t
\end{aligned}
\end{equation}
Then, by incorporating the original prediction equation \eqref{eq:approx_prior}, we have:

\begin{equation}\label{eq:qstar1}
\begin{aligned}
\hat{q}^{*}( \bm{x}_{t+1}, \bm{u} )
&= \int \int q(\bm{x}_{t+1}|\bm{x}_t, f_t) \, q(\bm{x}_t, \bar{\bm{u}}) \, \mathrm{d} \bm{x}_t \, \mathrm{d} f_t \\
&= \int \int q(\bm{x}_{t+1}|\bm{x}_t, f_t) \, p(f_t|\bm{u}) \, q(\bm{x}_t, \bm{u}) \, \mathrm{d} \bm{x}_t \, \mathrm{d} f_t \\
&= \int q(\bm{x}_{t+1}|\bm{x}_t, \bm{u}) \, q(\bm{x}_t, \bm{u}) \, \mathrm{d} \bm{x}_t 
\end{aligned}
\end{equation}
where $q(\bm{x}_{t+1}|\bm{x}_t, \bm{u})$ represents a new transition model, whose specific expression is:

\begin{equation}\label{eq:p_tran1}
\begin{aligned}
& q(\bm{x}_{t+1}|\bm{x}_t, \bm{u}) \\
&= \int q(\bm{x}_{t+1}|\bm{x}_t, f_t) \, p(f_t|\bm{u}) \, \mathrm{d} f_t \\
&= \int q(\bm{x}_{t+1}|\bm{x}_t, f_t) \, \mathcal{N}(f_t |\bm{k}_{tu} \bm{u}, \gamma) \, \mathrm{d} f_t \\
&= \mathcal{N} \Big(\bm{x}_{t+1} \Big| \bm{F}_t + \bm{A}_x (\bm{x}_t - \bm{\mu}_t) 
        + \bm{A}_f \bm{k}_{tu}(\bm{u} - \bm{m}_u), \tilde{\bm{\Sigma}}_f \Big)
\end{aligned}
\end{equation}
where $\gamma = \bm{K}_{tt} - \bm{K}_{tu} \bm{K}_{uu}^{-1} \bm{K}_{ut}$ and $\tilde{\bm{\Sigma}}_f = \bm{A}_f \gamma \bm{A}_f^T + \bm{\Sigma}_p$.

Combining \eqref{eq:qstar1} and \eqref{eq:p_tran1}, the prediction equation \eqref{eq:pred1} can be obtained.


\section{Derivation Details of GP Hyperparameters Optimization}\label{app:hp}

This section derives the moments of the approximate joint distribution $q_{\mathrm{new}}(\bm{x}_t, \bm{f})$ in \eqref{eq:qnew}, along with the optimizing objective for GP hyperparameters. 

Firstly, to derive the moments of the approximate joint distribution $q_{\mathrm{new}}(\bm{x}_t, \bm{f})$, we can utilize the following result:

\begin{equation}\label{eq:pno}
\begin{aligned}
\dfrac{p(\bm{u};\bm{\theta}_{\mathrm{new}})} {p(\bm{u};\bm{\theta}_{\mathrm{old}})}
&=  \dfrac{\mathcal{N}(\bm{u}|\bm{0}, \bm{K}_{uu}^{\mathrm{new}})} 
{\mathcal{N}(\bm{u}|\bm{0}, \bm{K}_{uu}^{\mathrm{old}}) } \\
&= Z \mathcal{N}(\bm{u}|\bm{0}, \Delta \bm{K}) \\
&=  Z \mathcal{N}(\bm{0}|\bm{u}, \Delta \bm{K})
\end{aligned}
\end{equation}
where 

\begin{equation}
\begin{aligned}
\Delta \bm{K} &= [(\bm{K}_{uu}^{\mathrm{new}})^{-1} - (\bm{K}_{uu}^{\mathrm{old}})^{-1}]^{-1}  \\ 
Z &\propto \dfrac{\vert \bm{K}_{uu}^{\mathrm{old}} \vert^{1/2}  \vert \Delta \bm{K} \vert^{1/2}}{\vert \bm{K}_{uu}^{\mathrm{new}} \vert^{1/2}}
\end{aligned}
\end{equation}
Note that, the above equations hold only in a mathematical sense and does not have probabilistic meaning, as \(\Delta \bm{K}\) may not be positive definite. Combining \eqref{eq:qnew} and \eqref{eq:pno}, we have:

\begin{equation}
q_{\mathrm{new}}(\bm{x}_t, \bm{u}) \propto \mathcal{N}(\bm{0}|\bm{u}, \Delta \bm{K}) q_{\mathrm{old}}(\bm{x}_t, \bm{u})
\end{equation}
whose moments can be obtained by the correction equation of Kalman filter:

\begin{equation}
\begin{aligned}
&\tilde{\bm{H}} =  \begin{bmatrix} \bm{0} & \bm{I}_{n_u} \\\end{bmatrix} \\
&\tilde{\bm{G}} = \bm{\Sigma}_{t}^{\text{old}} \tilde{\bm{H}}^T \left(\bm{S}_{uu}^{\text{old}} + \Delta \bm{K} \right)^{-1} \\
&\bm{\xi}_{t}^{\text{new}} = \bm{\xi}_{t}^{\text{old}} - \tilde{\bm{G}} \bm{\xi}_{t}^{\text{old}} \\
&\bm{\Sigma}_{t}^{\text{new}} = \bm{\Sigma}_{t}^{\text{old}}  - \tilde{\bm{G}} \tilde{\bm{H}} \bm{\Sigma}_{t}^{\text{old}}
\end{aligned}
\end{equation}
where $\bm{\xi}_t$ and $\bm{\Sigma}_t$ represents the mean and covariance of the augmented state $\bm{X}_t = [\bm{x}_t^T, \bm{u}^T]^T$. 

Secondly, in order to optimize the GP hyperparameters, considering \eqref{eq:py_new} and \eqref{eq:pno}, we can maximize:

\begin{equation}
\begin{aligned}
& \log \dfrac{p(\bm{y}_{1:t};\bm{\theta}_{\mathrm{new}})} 
{p(\bm{y}_{1:t}; \bm{\theta}_{\mathrm{old}})} \\
&= \log \int  Z \mathcal{N}(\bm{0}|\bm{u}, \Delta \bm{K}) q_{\mathrm{old}}(\bm{x}_t, \bm{u}) \, \mathrm{d} \bm{x}_t \, \mathrm{d} \bm{u} \\
&= \log \mathcal{N}(\bm{0}|\bm{m}_{u}^{\text{old}}, \Delta \bm{K} + \bm{S}_{uu}^{\text{old}}) + \log Z
\end{aligned}
\end{equation}

Then, by dropping some $\bm{\theta}_{\mathrm{new}}$-irrelevant terms, the optimization objective can be transformed to minimize:

\begin{equation}
\begin{aligned}
\mathcal L &= \mathcal L_1 + \mathcal L_2
\end{aligned}
\end{equation}
where

\begin{equation}
\begin{aligned}
\mathcal L_1 &= (\bm{m}_{u}^{\text{old}})^T (\bm{S}_{uu}^{\text{old}} + \Delta \bm{K})^{-1} \bm{m}_{u}^{\text{old}} \\
\mathcal L_2 &= 
 -\log \vert \Delta \bm{K} \vert + \log \vert \bm{K}_{uu}^{\mathrm{new}} \vert 
+ \log \vert \Delta \bm{K} + \bm{S}_{uu}^{\text{old}} \vert \\
&=  \log \vert \bm{K}_{uu}^{\mathrm{new}} \vert 
+ \log \vert \bm{I}_{n_u} + \Delta \bm{K}^{-1} \bm{S}_{uu}^{\text{old}} \vert \\
&= \log \vert \bm{K}_{uu}^{\mathrm{new}} + [ \bm{I}_{n_u} -\bm{K}_{uu}^{\mathrm{new}} (\bm{K}_{uu}^{\mathrm{old}})^{-1} ] \bm{S}_{uu}^{\text{old}} \vert
\end{aligned}
\end{equation}

Here, we avoid the appearance of $\Delta \bm{K}$ in the loss function through some operations, because when $\bm{K}_{uu}^{\mathrm{new}} = \bm{K}_{uu}^{\mathrm{old}}$, $\Delta \bm{K}$ is not well-defined.

\section{Stable Implementation Method}\label{app:stable}

To solve the numerical stability problem in RGPSSM, this section develops a stable implementation method based on Cholesky decomposition. Specifically, the covariance \( \bm{\Sigma}_t \) of the augmented state \( \bm{X}_t \) is decomposed as \( \bm{\Sigma}_t = \bm{L}_t \bm{L}_t^T \), where \( \bm{L}_t \) is a lower triangular matrix, known as the Cholesky factor. Based on this decomposition, the evolution of the covariance \( \bm{\Sigma}_t \) is replaced by the evolution of the Cholesky factor, which can reduce the accumulation of computational rounding errors. To implement this method, we must modify the moment matching equations within the prediction and correction steps, as well as during the discarding of inducing points and hyperparameter optimization.

For the prediction step, there exist two cases: not adding points and adding points. Firstly, for the prediction equation without adding points \eqref{eq:pred1}, the covariance update can be rewritten by:

\begin{equation}\label{eq:qr0}
\begin{aligned}
\bm{\Sigma}_{t+1}^- &= \begin{bmatrix}
\bm{\Phi}_t \bm{L}_t & \bm{D}_{p} \\
\end{bmatrix} 
\begin{bmatrix}
\bm{L}_t^T \bm{\Phi}_t ^T \\ \bm{D}_{p}^T \\
\end{bmatrix}
\end{aligned}
\end{equation}
where \( \bm{D}_p \) is the Cholesky factor of the process noise covariance \( \bm{\Sigma}_{p,\bm{X}} \). Then, through the following QR decomposition:

\begin{equation}\label{eq:qr1}
\begin{aligned}
\begin{bmatrix}
\bm{L}_t^T \bm{\Phi}_t ^T \\ \bm{D}_{p}^T \\
\end{bmatrix} = \bm{O} \bm{R}
\end{aligned}
\end{equation}
where \( \bm{O} \) is an orthogonal matrix and \( \bm{R} \) is an upper triangular matrix, we can obtain the Cholesky factor of the predicted covariance \( \bm{\Sigma}^-_{t+1} \), namely, \( \bm{L}_{t+1}^- = \bm{R}^T \).

Then, for the case of prediction with adding points, it involves one additional step: evaluating the Cholesky factor \( \bar{\bm{L}}_t \) of the augmented covariance \( \bar{\bm{\Sigma}}_t \) after adding a point. According to the definition, the augmented covariance \( \bar{\bm{\Sigma}}_t \) can be expressed as follows:
\begin{equation}\label{eq:chol_pred1}
\bar{\bm{\Sigma}}_t = \begin{bmatrix}
\bm{\Sigma}_{t} & \bm{\zeta}_t \\
\bm{\zeta}_t^T & \bm{S}_{tt} \\
\end{bmatrix}, \quad
\bm{\zeta}_t = \begin{bmatrix}
\bm{V}_{xt} \\
\bm{S}_{ut} \\
\end{bmatrix}
\end{equation}
In view of this expression, we can define:

\begin{equation}
\begin{aligned}
\bar{\bm{L}}_t = \begin{bmatrix}
\bm{L}_t & \bm{0} \\
\bm{\alpha} & \bm{\beta} \\
\end{bmatrix}
\end{aligned}
\end{equation}
and correspondingly we have:

\begin{equation}\label{eq:chol_pred2}
\begin{aligned}
\bar{\bm{\Sigma}}_t = \begin{bmatrix}
\bm{L}_t \bm{L}_t^T & \bm{L}_t \bm{\alpha}^T \\
\bm{\alpha}^T \bm{L}_t^T & \bm{\alpha}\bm{\alpha}^T + \bm{\beta}\bm{\beta}^T  \\
\end{bmatrix}
\end{aligned}
\end{equation}
Combining \eqref{eq:chol_pred1} and \eqref{eq:chol_pred2}, we can obtain \( \bm{\alpha} \) by solving the linear equation \( \bm{L}_t \bm{\alpha}^T = \bm{\zeta}_t \), and then compute \( \bm{\beta} = \mathrm{cholesky}\left(\bm{S}_{tt} - \bm{\alpha}\bm{\alpha}^T \right) \), thus obtaining the Cholesky factor \( \bar{\bm{L}}_t \). 
Then, similar operations as in \eqref{eq:qr0} and \eqref{eq:qr1} can be implemented to evaluate the predicted Cholesky factor.

For the correction step, given \eqref{eq:corr}, the update equation for covariance can be rewritten as:

\begin{equation}\label{eq:chol_post}
\begin{aligned}
\bm{\Sigma}_{t+1} &= \bm{\Sigma}_{t+1}^- - \bm{\Sigma}_{t+1}^- \bm{H}^T (\bm{C} \bm{P}_{t+1}^- \bm{C}^T + \bm{\Sigma}_m)^{-1} \bm{H} \bm{\Sigma}_{t+1}^- \\
&= \bm{\Sigma}_{t+1}^- - \bm{\Sigma}_{t+1}^- \bm{H}^T (\bm{\rho} \bm{\rho}^T)^{-1} \bm{H} \bm{\Sigma}_{t+1}^- \\
&= \bm{L}_{t+1}^{-} \bm{L}_{t+1}^{-T} - \bm{\eta} \bm{\eta}^T \\
\end{aligned}
\end{equation}
where \( \bm{\rho} = \mathrm{cholesky}(\bm{C} \bm{P}_{t+1}^- \bm{C}^T + \bm{\Sigma}_m) \) and \( \bm{\eta} = \bm{\Sigma}_{t+1}^{-} \bm{H}^T (\bm{\rho}^{-1})^T \). According to \eqref{eq:chol_post}, we can compute the Cholesky factor \( \bm{L}_{t+1} \) of the posterior covariance \( \bm{\Sigma}_{t+1} \) using the Cholesky downdate algorithm (see Section 6.5.4 of \cite{golubMatrixComputations2013}), that is, \( \bm{L}_{t+1} = \mathrm{choldowndate}(\bm{L}_{t+1}^-, \bm{\eta}) \). The downdate algorithm is more efficient, with quadratic complexity compared to the cubic complexity of direct decomposition.

For discarding an inducing point, to evaluate the updated Cholesky factor, we can express the original Cholesky factor before discarding as follows:

\begin{equation}\label{eq:ABC}
\begin{aligned}
\bm{L}_{t} = \begin{bmatrix}
\bm{A} & \bm{0} & \bm{0} \\
\bm{a} & \bm{b} & \bm{0} \\
\bm{B} & \bm{c} & \bm{C} 
\end{bmatrix}
\end{aligned}
\end{equation} 
Here, we reuse the notation \( \bm{A}, \bm{B}, \bm{C}, \bm{a}, \bm{b}, \bm{c} \), where \( \bm{a}, \bm{b}, \bm{c} \) denote the blocks corresponding to the discarded point \( u_{\mathrm{d}} \). Given \eqref{eq:res_proj}, the updated covariance \( \bm{\Sigma}_t^\prime \) is the original one with the \( i_{\mathrm{d}} \)th row and \( i_{\mathrm{d}} \)th column removed, which can be given using the expression in \eqref{eq:ABC}:

\begin{equation}
\begin{aligned}
\bm{\Sigma}_t^\prime = \begin{bmatrix}
\bm{A} \bm{A}^T & \bm{A} \bm{B}^T \\
\bm{B} \bm{A}^T & \bm{B} \bm{B}^T + \bm{c} \bm{c}^T + \bm{C} \bm{C}^T \\
\end{bmatrix}
\end{aligned}
\end{equation}
In view of this expression, the updated Cholesky factor $\bm{L}_t^\prime$ can be given by:

\begin{equation}\label{eq:chol_discard}
\begin{aligned}
\bm{L}_t^\prime = \begin{bmatrix}
\bm{A} & \bm{0} \\
\bm{B} & \mathrm{cholupdate}(\bm{C}, \bm{c}) \\
\end{bmatrix}
\end{aligned}
\end{equation}
where $\mathrm{cholupdate}(\cdot, \cdot)$ denotes the Cholesky update algorithm (see Section 6.5.4 of \cite{golubMatrixComputations2013}), namely a efficient method for evaluating the Cholesky factor of $(\bm{c} \bm{c}^T + \bm{C} \bm{C}^T)$.

For hyperparameter optimization, since $\Delta \bm{K}$ in \eqref{eq:update_hp} may not be positive definite as illustrated in Appendix \ref{app:hp}, we cannot update the Cholesky factor as in the correction step. Therefore, the updated Cholesky factor is evaluated by directly decomposing the updated covariance $\bm{\Sigma}_{t}^{\text{new}}$, which is obtained using \eqref{eq:update_hp}:

\begin{equation}\label{eq:chol_hp}
\begin{aligned}
    \bm{L}_{t}^{\text{new}} = \mathrm{cholesky}\left( \bm\Sigma_t^{\text{new}} \right)
\end{aligned}
\end{equation}

Overall, the Cholesky versions of the update equations in the prediction and correction steps, as well as for discarding inducing points and optimizing hyperparameters, have been derived, which makes the computations more compact, thereby enhancing numerical stability.

                                        
\end{document}